\documentclass[lettersize,journal]{IEEEtran}
\usepackage{amsmath,amsfonts}
\usepackage{algorithmic}
\usepackage{algorithm}
\usepackage{array}
\usepackage[caption=false,font=normalsize,labelfont=sf,textfont=sf]{subfig}
\usepackage{textcomp}
\usepackage{stfloats}
\usepackage{url}
\usepackage{verbatim}
\usepackage{graphicx}
\usepackage{cite}
\usepackage{amssymb}
\usepackage{amsmath}
\usepackage{multirow}
\usepackage{booktabs}
\usepackage{ragged2e}
\usepackage{diagbox} 
\usepackage{bm} 
\usepackage[colorlinks,linkcolor=red,anchorcolor=blue,citecolor=green]{hyperref}
\makeatletter

\newcommand{\Rmnum}[1]{\expandafter\@slowromancap\romannumeral #1@}
\makeatother
\begin{document}

\title{Quality Evaluation of Arbitrary Style Transfer: Subjective Study and Objective Metric}

\author{Hangwei Chen, Feng Shao,~\IEEEmembership{Member,~IEEE}, Xiongli Chai, Yuese Gu, Qiuping Jiang,~\IEEEmembership{Member,~IEEE}, \\Xiangchao Meng,~\IEEEmembership{Member,~IEEE}, Yo-Sung Ho,~\IEEEmembership{Fellow,~IEEE}
\thanks{This work was supported by the Natural Science Foundation of China (grant 62071261,  62076013, 62021003, 62271277), and Zhejiang Province Natural Science Foundation of China (grant R18F010008, LR22F020002). (\emph{Corresponding author: Feng Shao.})}        
\thanks{Hangwei Chen, Feng Shao, Xiongli Chai,  Yuese Gu, Qiuping Jiang, and Xiangchao Meng are with the Faculty of Information Science and Engineering, Ningbo University, Ningbo 315211, China (e-mail: 1010075746@qq.com; shaofeng@nbu.edu.cn; 747866472@qq.com; 805682724@qq.com; jiangqiuping@nbu.edu.cn, mengxiangchao@nbu.edu.cn).}
\thanks{Yo-Sung Ho is with the School of Information and Communications, Gwangju Institute of Science and Technology (GIST), Gwangju 500-712, Korea (e-mail: hoyo@gist.ac.kr).}}

\markboth{IEEE TRANSACTIONS ON CIRCUITS AND SYSTEMS FOR VIDEO TECHNOLOGY}%
{Shell \MakeLowercase{\textit{et al.}}: A Sample Article Using IEEEtran.cls for IEEE Journals}


\maketitle

\begin{abstract}
Arbitrary neural style transfer is a vital topic with great research value and wide industrial application, which strives to render the structure of one image using the style of another. Recent researches have devoted great efforts on the task of arbitrary style transfer (AST) for improving the stylization quality. However, there are very few explorations about the quality evaluation of AST images, even it can potentially guide the design of different algorithms. In this paper, we first construct a new AST images quality assessment database (AST-IQAD), which consists 150 content-style image pairs and the corresponding 1200 stylized images produced by eight typical AST algorithms. Then, a subjective study is conducted on our AST-IQAD database, which obtains the subjective rating scores of all stylized images on the three subjective evaluations, i.e., content preservation (CP), style resemblance (SR), and overall vision (OV). To quantitatively measure the quality of AST image, we propose a new sparse representation-based method, which computes the quality according to the sparse feature similarity. Experimental results on our AST-IQAD have demonstrated the superiority of the proposed method. The dataset and source code will be released at \href{https://github.com/Hangwei-Chen/AST-IQAD-SRQE}{https://github.com/Hangwei-Chen/AST-IQAD-SRQE}
\end{abstract}


\section{Introduction}
\subsection{Background}
\begin{figure}[!t]
    \centering
    \includegraphics[width=3.5in]{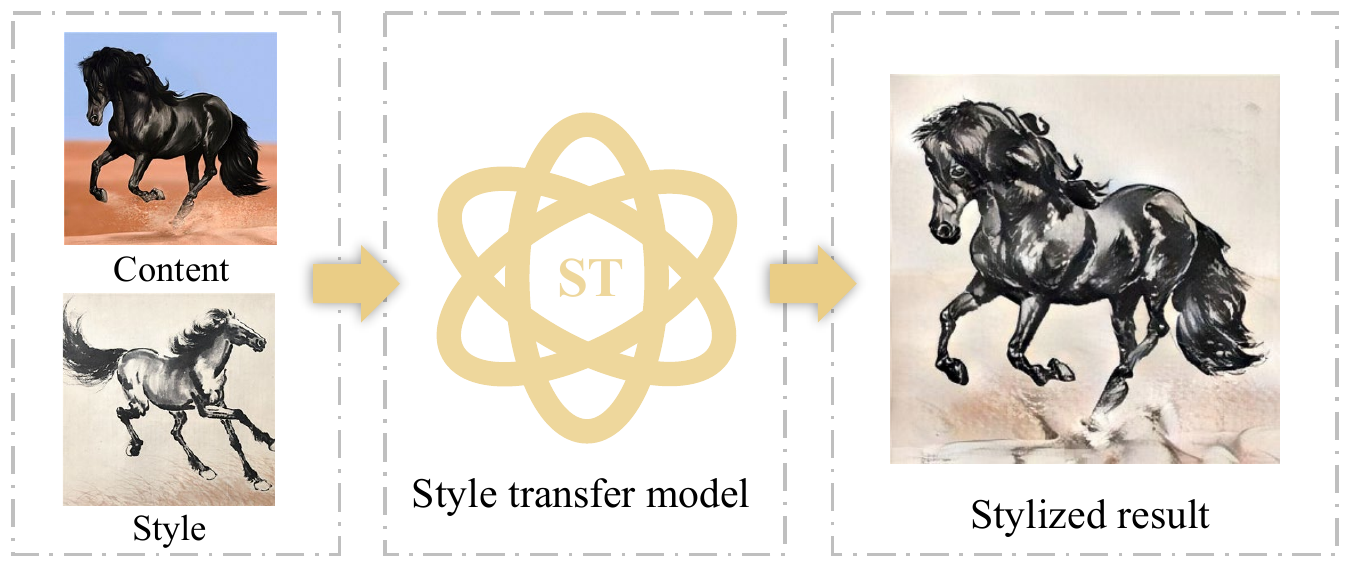}
    \caption{An example of style transfer model that generates a stylized image.}
    \label{fig1}
\end{figure}

\IEEEPARstart{S}{tyle} transfer is a process that strives to render natural images with particular style characteristics from one image (e.g., the style image) while synchronously maintaining the detailed structure information of the content image. Such unique technique not only builds a bridge between the computer vision and appealing artworks, but also gets rid of the dilemma that it would take a long time for a well-trained artist to draw an image in a special style \cite{ref1}. As shown by an example in Fig. \ref{fig1}, style transfer model can automatically generate a new stylized image based on the content and style of an image. Additionally, style transfer also plays an important role in many computer vision tasks, such as person re-identification \cite{ref2}, semantic segmentation \cite{ref3}, and image reconstruction \cite{ref4}.

As customary, style transfer is commonly cast as the study of the texture generation \cite{ref5}. Early works \cite{ref6} cope with the texture generation using local statistics or similarity measures on the pixel values. Recently, the field of Neural Style Transfer (NST) was ignited by the groundbreaking work of Gatys \emph{et al}. \cite{ref7}, which is the process of using Convolutional Neural Network (CNN) to perform image translation and stylization. Then, lots of follow-up studies were conducted on the NST algorithm based on the deep neural network in order to promote the transfer efficiency and generation effects. Meanwhile, NST also has rapidly evolved from single-style to infinite-style models, also known as Arbitrary Style Transfer (AST) \cite{ref8,ref9,ref10,ref11,ref12,ref13,ref14,ref15,ref16,ref17,ref18,ref19,ref20}. With the ability of utilizing one model to transfer arbitrary artistic style, AST has become a hot topic on computer vision, which could promote the creation of artworks and social communication. However, lacking of a good quantitative evaluation makes it difficult to measure the merits and drawbacks of the AST algorithms. It is therefore necessary to study and eventually evaluate the AST images.

\subsection{Style Transfer Techniques}
Mathematically, the style transfer model can be described as a translation process \cite{ref21}:
\begin{equation}
\label{deqn_ex1a}
I_{AB}\in B:\mathcal{M} _{A\longmapsto B} (I_{A} ).
\end{equation}
where $I_{A}$ is the input content image from a source domain \emph{A} to a target domain \emph{B}, $\mathcal{M}_{A\longmapsto B}$ is a mapping that generates image $\emph{I}_{AB}\in\emph{B}$ indistinguishable from style image $\emph{I}_{B}\in\emph{B}$ on the target domain given the input source image $\emph{I}_{A}\in\emph{A}$.

Following the growth of neural networks, numerous NST methods have been proposed to study the problem of style transfer. According to the optimization ways adopted in the NST task, the existing NST methods can be divided into two categories: Image-Optimization-Based Online Neural Methods \cite{ref8,ref9} and Model-Optimization-Based Offline Neural Methods \cite{ref12,ref13,ref14,ref15,ref16,ref17,ref18,ref19,ref20}. The former category could produce appealing stylized results through the iterative image optimization process, while the latter uses generated models with feed-forward networks to produce special style patterns. Particularly, considering the efficiency issue, it is more meaningful to design the Model-Optimization-Based Offline Neural Methods in practice.

The AST method is one of the Model-Optimization-Based Offline Neural Methods, which can accept an arbitrary artistic style as input and produce stylized results in a single feed-forward network once upon the model is trained. Thus, the AST has received substantial attention due to the increasing scientific and artistic values. In below, we review the details of the AST studies. For more comprehensive introduction, readers can refer to the survey \cite{ref1}.

\emph{1) Non-Parametric Methods:} The common idea of the non-parametric methods \cite{ref10,ref11} is to seek the similarities between the patches in content and style images and swap them. Chen \emph{et al}. \cite{ref10} realized AST for the first time that developed a Style-Swap operation to swap the feature patches of content images with the best matching style feature patches. Another work by Gu \emph{et al}. \cite{ref11} also proposed a patch-based method considering the matching of both global statistics and local patches. However, if the structures of content image and style image are largely different, these methods cannot efficiently protect the shape with unsatisfactory style patterns.

\emph{2) Parametric Methods:} The characteristic of these parametric methods \cite{ref12,ref13,ref14,ref15,ref16,ref17,ref18,ref19} is to optimize a target function that reflects the similarity between the input and stylized images. These parametric methods \cite{ref12,ref13,ref14}, utilize the Gram-based VGG perceptual loss to produce stylization with a few modifications. Li \emph{et al}. \cite{ref12} proposed a linear transform function (LST) from content and style features for stylization. Li \emph{et al}. \cite{ref13} performed a pair of feature transforms, whitening and coloring (WCT), for feature embedding within a pre-trained encoder-decoder module. Huang \emph{et al}. \cite{ref14} proposed a novel adaptive instance normalization (AdaIN) layer that adjusts the mean and variance of the content input to match the style input. Another promising trend in parametric methods is to integrate attention mechanism into the deep neural network. Yao \emph{et al}. \cite{ref15} first considered multi-strokes with self-attention mechanism. Park\emph{et al}. \cite{ref16} introduced Style-Attentional Network (SANet) to match content and style features for achieving good results with evident style patterns. Deng\emph{et al}. \cite{ref17} proposed a multi-adaptation module that takes the global content structure and local style patterns into account. In addition, Zhang \emph{et al}. \cite{ref18} introduced a multi-modal style transfer (MST) via efficient graph cuts algorithms, which explicitly considers the matching of semantic patterns in content and style images. Inspired by MST, Chen \emph{et al}. \cite{ref19} developed a structure-emphasized multimodal style transfer (SEMST) model, which can flexibly match the content and the style clusters based on the cluster center norm.
\subsection{Image Quality Assessment of AST}
\emph{1) Motivations:} Notwithstanding the current state-of-the-art methods have shown successful stream in style transfer, arbitrary style transfer image quality assessment (AST-IQA) has been a long-standing problem and relatively unexplored in the community. Nevertheless, with the exception of a few quantitative protocols \cite{ref22,ref23}, almost all researches evaluate the stylization quality in a qualitative way (e.g., by side-by-side subjective vision comparisons or different user studies), which suffers from the following limitations. First, the stylization examples displayed for qualitative comparison are limited in number and often carefully selected to favor the cases where the algorithm works well. In other words, the results of these presentations are not comprehensive enough. Second, the selected observers often lack sufficient experience and expertise in art, which makes qualitative evaluation less convincing. Thus, it is practical and necessary to propose a reliable metric to quantitatively assess the stylization quality.

\emph{2) Challenges:} Different from the traditional IQA tasks \cite{ref24,ref25,ref26,ref27,ref28,ref29} that usually focus on general distortions generated by various stimuli, the AST-IQA is closely related to aesthetics and poses serious challenges in both subjective and objective assessment. 

\textbf{Subjective assessment:} The first challenge is how to design and conduct a human subjective study that can obtain reliable ground truth labeling on a set of stylized images \cite{ref30}. To our best knowledge, there is no publicly available database for AST-IQA. The most related benchmarks are the non-photorealistic rendering (NPR) benchmarks \cite{ref31,ref32}, which are used for testing stylization algorithms without human opinion scores. Currently, there is no quality assessment standard for measuring the performance of style transfer, since the AST-IQA is a highly subjective task, e.g., different subjects tend to have various ideas towards the same stylized result, especially for the style evaluation.

\textbf{Objective assessment:} Once the subjective dataset is obtained, the next challenge is how to design a metric that can automatically evaluate the perceptual quality of the AST images closely consistent to human vision. In several image style transfer works \cite{ref21,ref33,ref34}, some Full-reference (FR) metrics (e.g., SSIM \cite{ref35}) have been used to evaluate the similarity between the structures of the content and stylized images. However, strictly speaking, evaluating the AST quality is not a classic FR-IQA task. Straightforwardly applying the traditional FR-IQA strategy to the field of style transfer is problematic, since the stylized image has different content detailed information with the source and style images, and the ground truth image is unavailable for the stylized image (i.e., completely different with fidelity evaluation in the tradition IQA tasks). Although the general No-reference (NR) metrics (e.g., NIQE \cite{ref28}, TCLT \cite{ref36} and BRISQUE \cite{ref37}) have made great progress in the tradition IQA tasks without ground truth, they are also not applicable to AST-IQA because stylized images are closely related to aesthetics rather than naturalness. In addition, it is necessary for AST-IQA to fully take the original information of content and style images into account. Recent works target to address these challenges using some objective metrics from the perspective of different quality factors. Yeh \emph{et al}. \cite{ref22} proposed a metric with two factors (i.e., effectiveness and coherence) where the former factor is a measure of the extent to which the style was transferred, and the latter is a measure of the extent to which the transferred image is decomposed into the content objects. Wang \emph{et al}. \cite{ref23} first decomposed the quality of style transfer into three quantifiable factors, i.e., the content fidelity (CF), global effects (GE) and local patterns (LP), which cover the main aspects considered by different types of existing NST methods. However, these metrics either focus on the limited factors of style transfer quality (e.g., lacking of fine-grained quality factors), or are simple in quality pooling, which cannot effectively match the aesthetic perception of human observers in practice.
\begin{figure}[!t]
    \centering
    \includegraphics[width=3.3in]{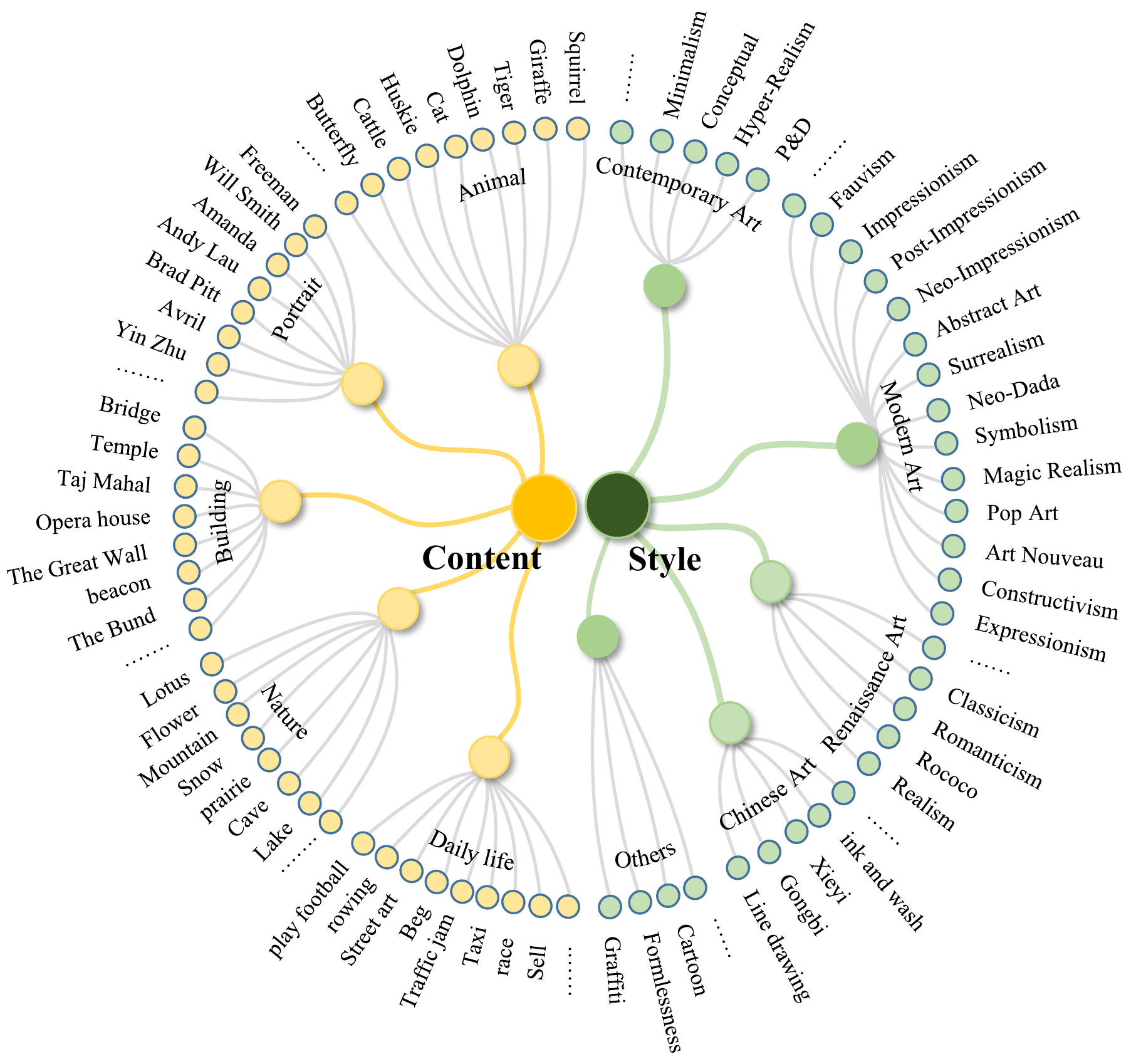}
    \caption{Taxonomic structure of our source images.}
    \label{fig2}
\end{figure}
\subsection{Overview of Our Work}
In this paper, to resolve the above challenges, we carry out an in-depth investigation for AST-IQA from both subjective and objective perspectives. With the popularization of art education, a majority of people with similar background can make similar perceptual judgments about some basic elements in artistic painting (e.g., color tone, brush stroke, distribution of objects, and contents). Benefitted from the above conditions, to address the challenge in subjective assessment, we decompose the quality of AST into three quality factors that are easier to understand, namely content preservation (CP), style resemblance (SR), and overall vision (OV). These quality factors are assigned own preference labels by participants according to the knowledge in style transfer, intuition in vision and feed-back from the surveys. To address the challenge in objective assessment, as suggested by the recommendation system \cite{ref38}, we regard the problem as a data-driven modeling of user preference \cite{ref39}, and conduct quantitative evaluation of AST-IQA using sparse representation to dig intrinsic representation for content and style images. To sum up, the major contributions of our work are summarized as follows:

1) To carry out in-depth studies on perceptual quality assessment of AST stylized images from both subjective and objective aspects, we build a new AST images database named AST-IQAD, which consists 150 content-style image pairs and the corresponding 1200 stylized images produced by eight typical AST algorithms. Each stylized image contains the subject-rated CP, SR and OV scores. To our knowledge, it is the first large-scale AST image database with human opinion scores. Therefore, it can serve as a benchmark to objectively evaluate the existing AST methods and potentially guide the design of different AST methods.

2) We proposed a new sparse representation-based image quality evaluation metric (SRQE) for AST-IQA, which can quantitatively evaluate the quality factors of CP, SR, and OV. To be more specific, in the training phase, we learn multi-scale style and content dictionaries to represent the style characteristic and structure of the stylized images. In the quality estimation phase, the sparse feature similarities are further exploited to compute the qualities of CP and SR respectively, and the OV quality is obtained by combining the SR and CP qualities. Extensive experiments are conducted on our AST-IQAD dataset and the experimental results demonstrate the proposed method can well evaluate the AST quality.

The rest of this paper is organized as follows. Section II illustrates the details of AST-IQAD. Section III introduces the proposed method in detail. The experimental results are shown and discussed in Section IV. Finally, conclusions are drawn in Section V.
\begin{figure}[!t]
    \centering
    \includegraphics[width=3.5in]{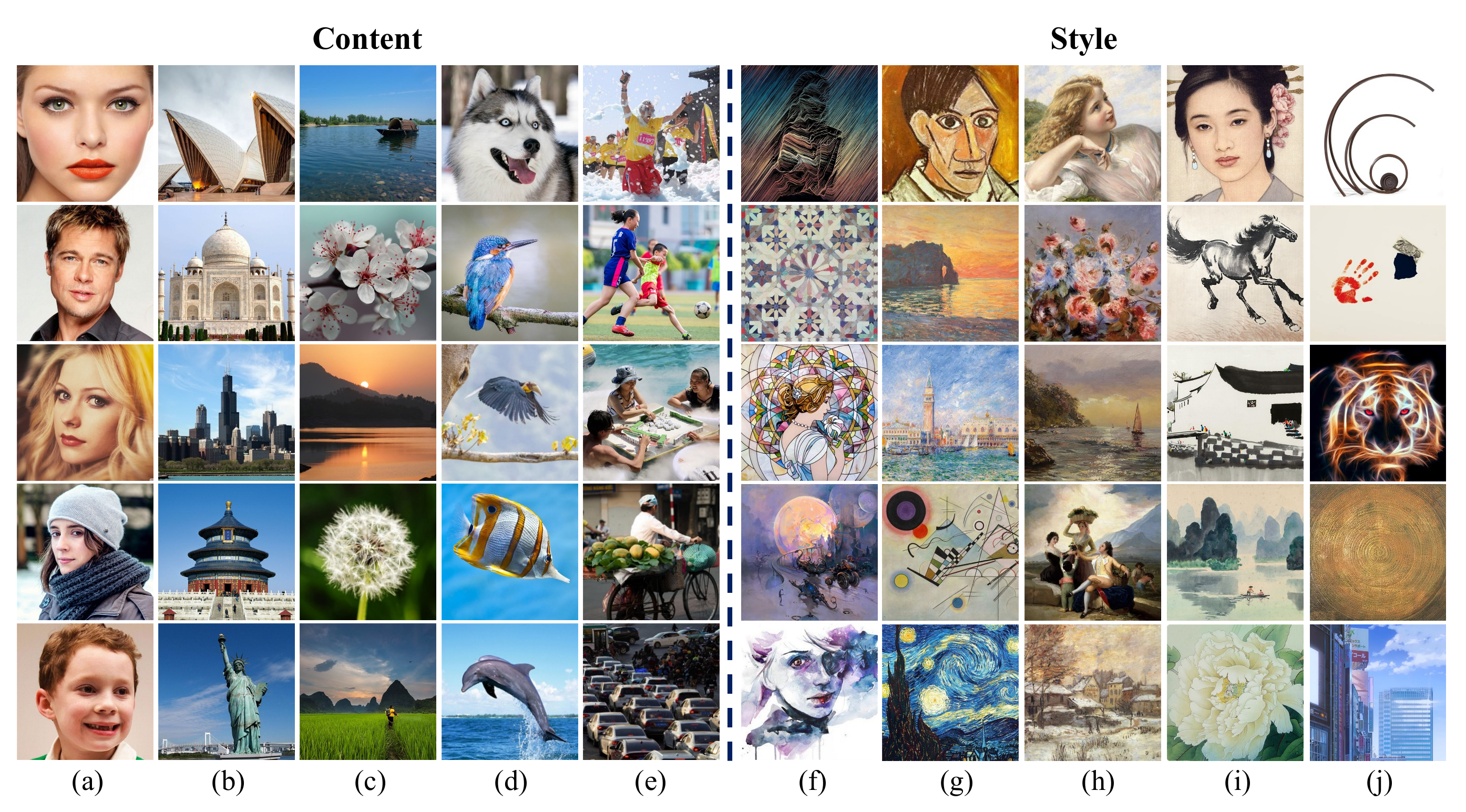}
    \caption{Examples of content and style images in the AST-IQAD database. (a) Portraits. (b) Building. (c) Nature. (d) Animal. (e) Daily life. (f) Contemporary art. (g) Modern art. (h) Renaissance art. (i) Chinese art. (j) Others.}
    \label{fig3}
\end{figure}

\section{AST-IQAD DATABASE}
To investigate quality assessment of AST images, we construct a new arbitrary style transfer database (AST-IQAD) for quality assessment, which includes 1200 stylized images generated by eight typical AST methods, and conduct a subjective quality evaluation study on the AST-IQAD database to capture the human option scores. To our knowledge, it is the first large-scale database for AST-IQA, and it can provide a better resource to evaluate and advance state-of-the-art style transfer algorithms. We will introduce the details of the AST-IQAD database in the following parts.
\subsection{Source Images}
Since the essence of style transfer is to migrate the color tone and stroke pattern from the source to target image while retaining the content structure information of the target image. To provide deeper and intuitive information, the selected source images should have clear and reasonable structures. Thus, we set up a hierarchical taxonomic system (shown in Fig. \ref{fig2}) for source image (the content images) and target images (the style images), respectively. Both content and style images are labeled with five categories.

\emph{1) Content Images:} We collect 75 high quality images with a resolution of $512\times512$ pixels from the NPRgeneral benchmark \cite{ref31} and other famous photography websites. According to the criteria of coverage \cite{ref31}, the content images are comprised of five categories (i.e., animal, portrait, building, nature, and daily life) with a wide range of characteristics (e.g., contrast, texture, edges and meaningful structures). Examples of the selected content images in the database are shown in Fig. \ref{fig3}.

\begin{figure*}[t]
    \centering
    \includegraphics[width=6.9in]{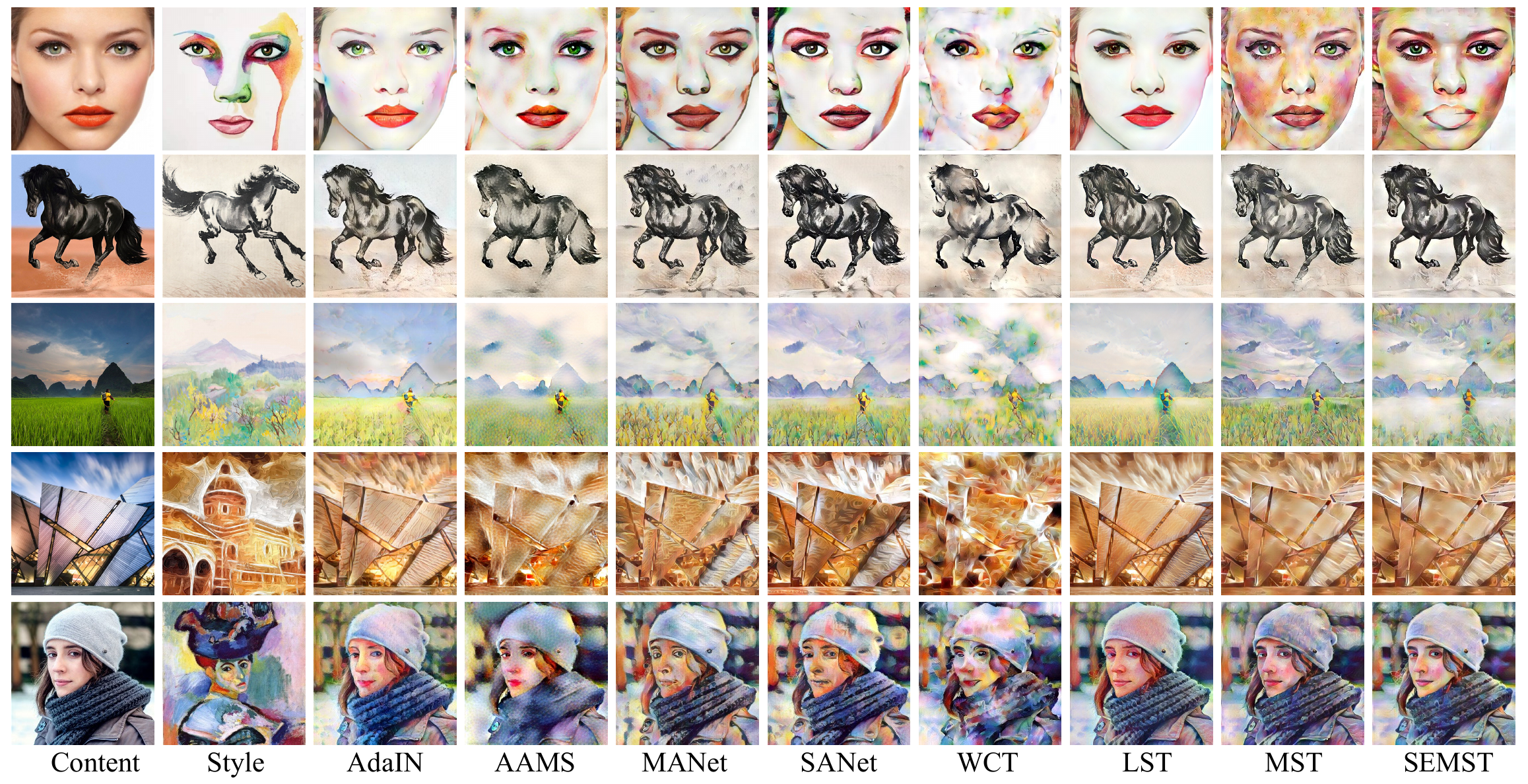}
    \caption{Examples of AST images produced by different algorithms.}
    \label{fig4}
\end{figure*}
\begin{figure}[t]
    \centering
    \includegraphics[width=3.3in]{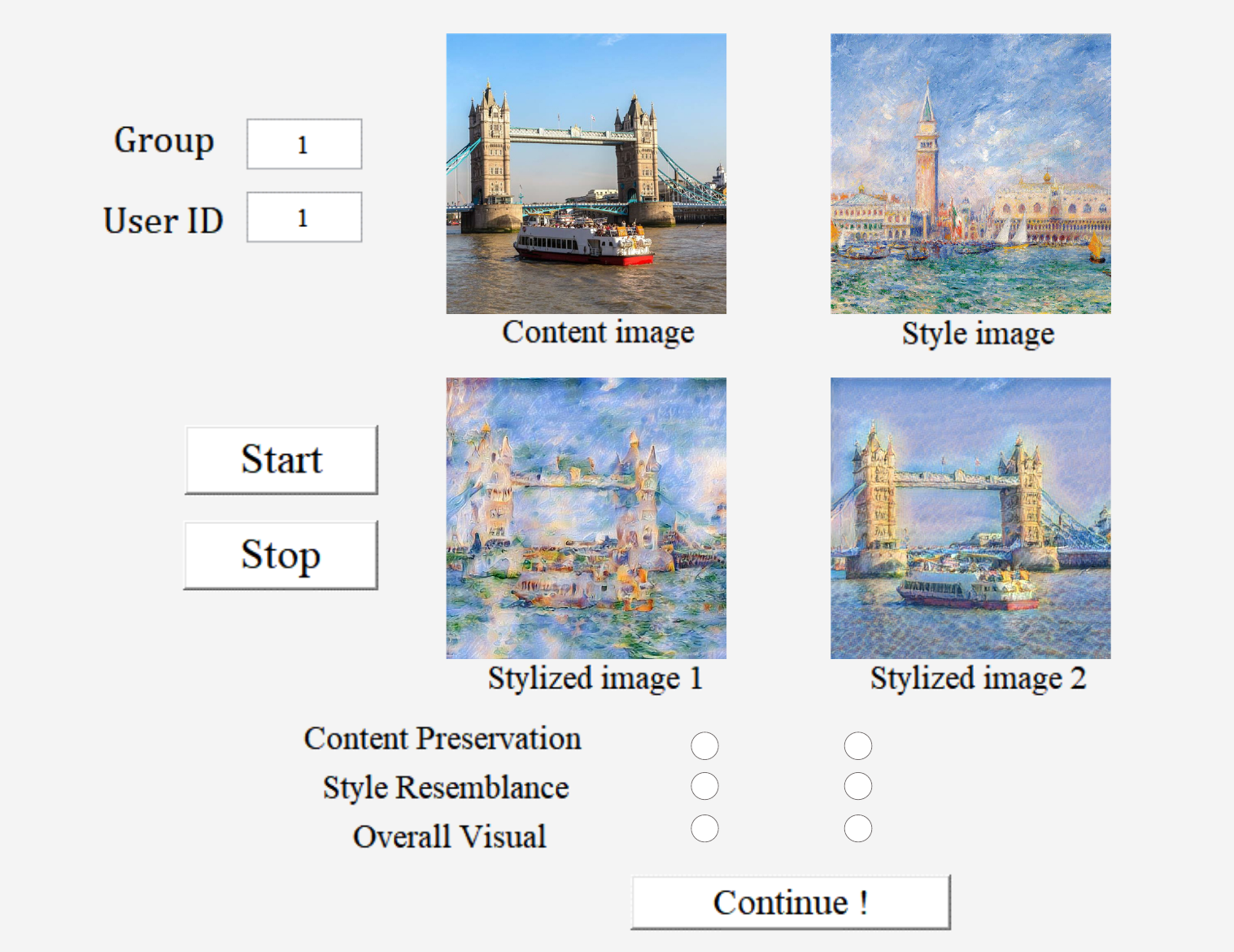}
    \caption{Subjective interface in the experiment.}
    \label{fig5}
\end{figure}
\begin{table}[t]
\begin{center}
\caption{Description of AST algorithms used in the database.}
\label{tab1}
\begin{tabular}{lll}
\toprule
Types                            & Methods        & Descriptions                                                                              \\ \midrule
\multirow{3}{*}{Gram-based}      & WCT   \cite{ref13}   & Whitening and coloring transforms                                                         \\  
                                 & AdaIN \cite{ref14} & Adaptive instance normalization                                                           \\ 
                                 & LST   \cite{ref12}   & Linear style transfer                                                                     \\ \midrule
\multirow{3}{*}{Attention-based} & AAMS  \cite{ref15}  & \begin{tabular}[c]{@{}l@{}}Attention-aware Multi-stroke \\ style transfer\end{tabular}    \\  
                                 & MANet \cite{ref17} & Multi-adaptation networks                                                                 \\ 
                                 & SANet \cite{ref16} & Style-attentional networks                                                                \\ \midrule
Graph-based                      & MST   \cite{ref18}   & \begin{tabular}[c]{@{}l@{}}Multimodal style transfer via \\ graph cuts\end{tabular}       \\ \midrule
Cluster-based                    & SEMST \cite{ref19} & \begin{tabular}[c]{@{}l@{}}Structure-emphasized multimodal \\ style transfer\end{tabular} \\ \bottomrule
\end{tabular}
\end{center}
\end{table}

\emph{2) Style Images:} We select 126 style images from some artwork websites and WikiArt. WikiArt is the largest art encyclopedia in the visual arts from all over the world. Since these original style images are not provided with high resolution, all style images are set to the same resolution of $512\times512$ in line with the content images. These style images cover five categories including contemporary art, modern art, renaissance art, Chinese art, and others. Examples of style images in the AST-IQAD dataset are also shown in Fig. \ref{fig3}.

\emph{3) Content-Style Image Pairs:} Pairing content images with appropriate and diverse style images can make the AST results more aesthetically pleasing. In our work, we provide two ‘Paired’ and ‘Unpaired’ mechanisms \cite{ref40} for each content image, in which ‘Paired’ means that the content and the style images are semantically consistent (e.g., the same source of birds), while ‘Unpaired’ means that the content and the style images are the representations of different sources (e.g., the style images may be regular patterns or texture decorations). In total, we use 75 content images and 126 style images (including reused style images) to generate 150 content-style image pairs. Then, eight AST algorithms are conducted on the content-style image pairs to generate the AST images. More information of the image pairs can be found in our database.

\subsection{Arbitrary Style Transfer Algorithms}
Different with traditional IQA databases that stimulated with different distortion stimuli, the testing images in our databases are generated from different AST methods. The eight representative AST methods used in the database are listed in Table \ref{tab1}, including AAMS \cite{ref15}, AdaIN \cite{ref14}, MANet \cite{ref17}, LST \cite{ref12}, WCT \cite{ref13}, MST \cite{ref18}, SEMST \cite{ref19}, SANet \cite{ref16}. These algorithms cover a wide variety of techniques, including Gram-based, Attention-based, Graph-based and Cluster-based methods. As a result, we can obtain 1200 style-transferred images from 150 content-style image pairs. As shown by the examples of style-transferred images in Fig. \ref{fig4}, we have the following observations: 1) In the Gram-based methods, LST \cite{ref12} and AdaIN \cite{ref14} can well preserve the content information but may suffer from wash-out artifacts. On the contrary, WCT \cite{ref13} is impressive in color and texture making the “painting taste” more intense while fails to preserve the main content structures. 2) The attention-based methods have distinct content structures and rich style patterns but may produce unpleasing visual artifacts. For example, SANet \cite{ref16} and MANet \cite{ref17} methods produce unpleasing eye-like artifacts, and AAMS \cite{ref15} introduces imperceptible dot-wise artifacts. 3) The stylized results of MST \cite{ref18} and SEMST \cite{ref19} are similar and produce both visible content and proper stylization.
\begin{figure}[]
    \centering
    \includegraphics[width=3.3in]{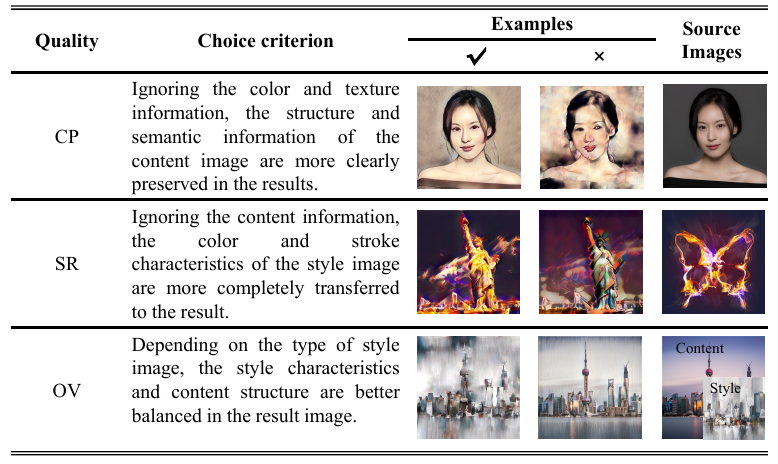}
    \caption{Preference criteria for the AST-IQA task.}
    \label{fig6}
\end{figure}

\subsection{Human Subjective Study}
Due to the different rating standards across different ob-servers and the influence of visual content \cite{ref41}, the subjective quality scores evaluated by absolute category rating are imprecise, biased, and inconsistent, while the preference label, representing the relative quality of two images, is generally precise and consistent for the task. For this consideration, we adopt the pairwise comparison (PC) approach which aims to provide a binary preference label between a pair of stylized images.

The experiment was carried out in an indoor laboratory with low ambient illumination calibrating in accordance with the ITU-R BT.500-13 recommendations \cite{ref42}. The subjective software interface (as shown in Fig. \ref{fig5}) is displayed at a 23-inch true color (32bits) LCD monitor with the screen resolution of $1920\times1080$ pixels. The ratio of background illumination behind the display to the image peak luminance is 0.15. The viewing distance is approximately six times the image height. In the interface specific to the AST task, participants are simultanesously shown two stylized images along with a style-content image pair, and are required to vote on three preferences for content preservation (CP), style resemblance (SR), and overall vision (OV). The detailed descriptions of preference criteria are summarized in Fig. \ref{fig6}. 

A total number of 45 subjects aged from 18 to 30, including experts and students from the Faculty of Art, and under-graduate students with experience in image processing, were participated in the subjective study. For each content-style image pair producing eight AST results, we have 28 pairwise comparisons in the subjective ranking study. In total, 4200 pairwise comparisons from 150 content-style pairs are involved in the subjective study. To reduce the possible fatigue effect, we divide the experiment into three sub-sessions (each sub-session contains 15 subjects), in which each participant would take part in one sub-session completing 1400 PC voting. The content images presented in each sub-session included most of the scene types, and were displayed in a random order to reduce the possible memory interference. After completing every 200 pair comparisons, each subject was encouraged to look away to relax their eyes and asked to rest for about 15 minutes. Each participant took about six hours (including rest periods) to complete the whole subjective experiment. As a result, we can get 63,000 votes on each subjective evaluation.

\subsection{Subjective Data Analysis}
\emph{1) Global Ranking of AST Algorithms:} To derive global ranking of the AST algorithms from the corresponding PC results, we adopt the Bradley-Terry \cite{ref43} model to estimate the subjective score for each algorithm. The probability that the \emph{i}-th method is favored over the \emph{j}-th method is defined as:
\begin{equation}
\label{deqn_ex1a}
P(i\succ j)=\frac{{{e}^{{{u}_{i}}}}}{{{e}^{{{u}_{i}}}}+{{e}^{{{u}_{j}}}}}
\end{equation}
where $\emph{u}_i$ and $\emph{u}_j$ are subjective scores prefered for the $\emph{i}$-th and the $\emph{j}$-th methods, respectively. Then, the negative log-likelihood for the B-T scores $\textbf{u}\in {{\mathbb{R}}^{{{n}_{ast}}}}$, where $\emph{n}_{ast}$ is the number of AST algorithms, can be jointly expressed as:
\begin{equation}
\mathsf{\mathcal{L}}(\mathbf{u} )=-\log \left( \prod\limits_{i=1}^{{{n}_{ast}}}{\prod\limits_{j=1,j\ne i}^{{{n}_{ast}}}{P{{(i\succ j)}^{{{\mathsf{\mathcal{W}}}_{ij}}}}}} \right)
\end{equation}
where ${{\mathsf{\mathcal{W}}}_{ij}}$ is the (\emph{i}, \emph{j})-th element in the winning matrix $\mathcal{W}\in {{\mathbb{Z}}^{{{n}_{ast}}\times {{n}_{ast}}}}$, representing the number of times that the \emph{i}-th method is preferred over the \emph{j}-th method. By setting the derivative of   $\mathsf{\mathcal{L}}(\mathbf{u})$ in Eq. (3) to zero to solve the optimization problem \cite{ref44}, the final B-T scores \textbf{u} are obtained via zero mean normalization, served as the ground truth subjective rating scores.
\begin{figure}[t]
    \centering
    \includegraphics[width=3.3in]{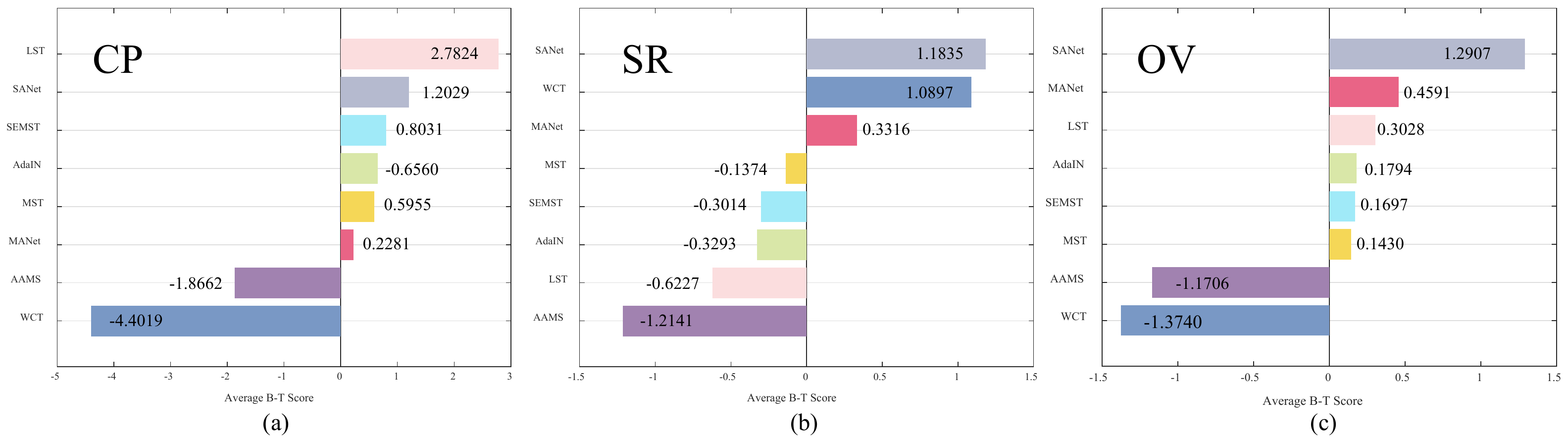}
    \caption{Average B-T scores of different AST algorithms at each subjective evaluation.}
    \label{fig7}
\end{figure}
\begin{figure}[t]
    \centering
    \includegraphics[width=3.3in]{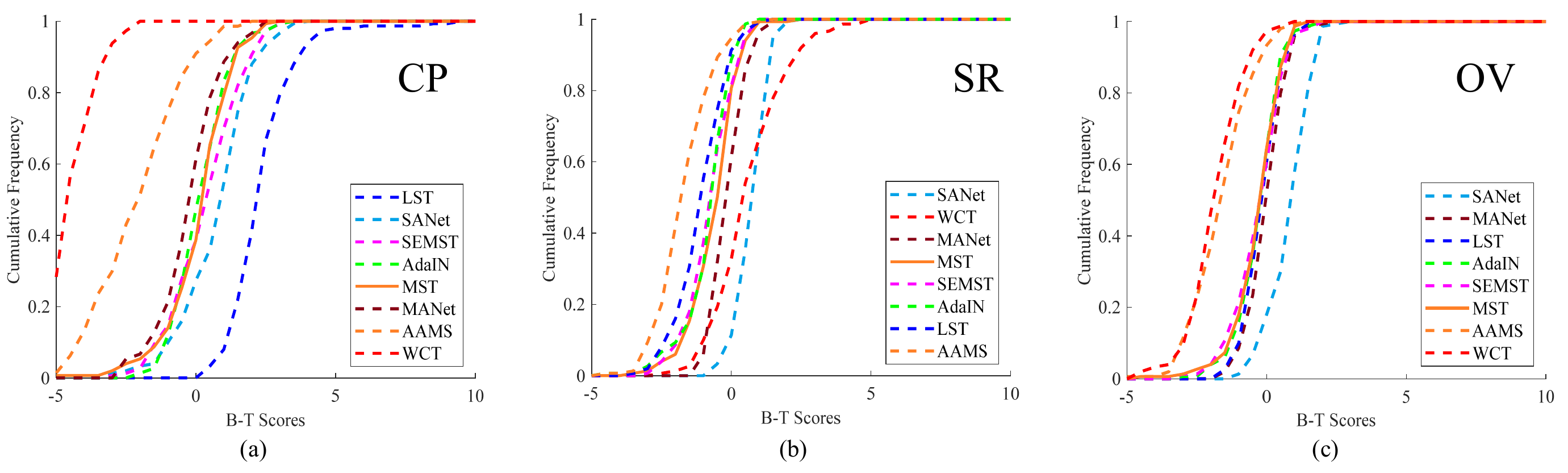}
    \caption{Cumulative probability distribution curves of B-T scores at each subjective evaluation.}
    \label{fig8}
\end{figure}
\begin{figure}[t]
    \centering
    \includegraphics[width=3.3in]{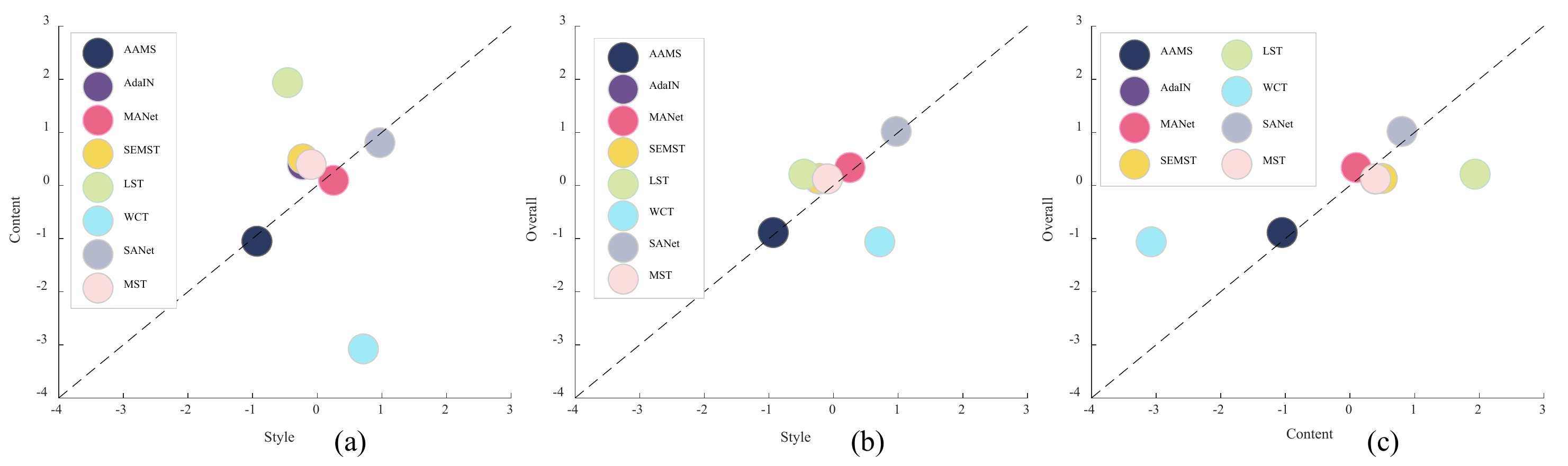}
    \caption{Correlation of B-T scores between the subjective evaluations of CP and SR, SR and OV, and CP and OV.}
    \label{fig9}
\end{figure}

We can get a B-T score for each AST result in each sub-session by applying the B-T model. Fig. \ref{fig7} shows the average B-T scores of different AST methods for three subjective evaluations (i.e., CP, SR, and OV). A higher B-T score indicates a better performance. From the figure, some interesting observations could be drawn: 1) The B-T scores of different AST algorithms show various trends on the subjective evaluations of CP, SR and OV, which indicate that different methods have specific advantages. 2) For the CP evaluation, LST \cite{ref12}, which receives the best ranking, has shown significant advantage in maintaining structure information over other methods by a large margin. WCT \cite{ref13} performs worst in the CP test due to treating diverse image regions in an indiscriminate way. 3) For the SR evaluation, SANet \cite{ref16} performs best on average, attributed to the attention mechanism that generates more local style details. However, AAMS \cite{ref15}, which is also based on self-attention mechanism, does not perform well because the multi-stroke pattern generates imperceptible dot-wise artifacts. In addition, WCT \cite{ref13} shows the competitive advantage, which indicates the effectiveness of whitening and coloring transformations. 4) For the OV evaluation, attention-based methods (e.g., SANet \cite{ref16} and MANet \cite{ref17}) rank top-2 on performance. It is not surprising because attention-based algorithms pay more attention to those feature-similar areas in the style image for stylizing a content image region. Furthermore, we plot the cumulative probability distribution curves of the B-T scores obtained from all AST results in Fig. \ref{fig8}. The AST method corresponding to the rightmost curve performs better because it accumulates higher B-T scores.

\emph{2) Correlation of B-T Scores:} This part analyzes the correlation of B-T scores between the subjective evaluation of CP and SR, SR and OV, and CP and OV, as shown in Fig. \ref{fig9}. Taking SR and CP as an example, the method above (below) the diagonal indicates better performance in CP (SR), and vice versa. From the Fig. \ref{fig9} (a), it is observed that the SANet \cite{ref16} method has achieved a better consistency in SR and CP evaluation, and the WCT \cite{ref13} method shows excellent performance in SR but fails to sufficiently maintain the content structure. In addition, the correlation phenomena observed in Fig. \ref{fig9} (b) and (c) also reveal the complex relationship between OV and CP (or SR).

\emph{3) Convergence Analysis:} To demonstrate that the scale of the subjective study is large enough to support performance evaluation, we further analyze the convergence from the perspectives of the number of subject votes and images pairs.

\begin{figure}[]
    \centering
    \includegraphics[width=3.3in]{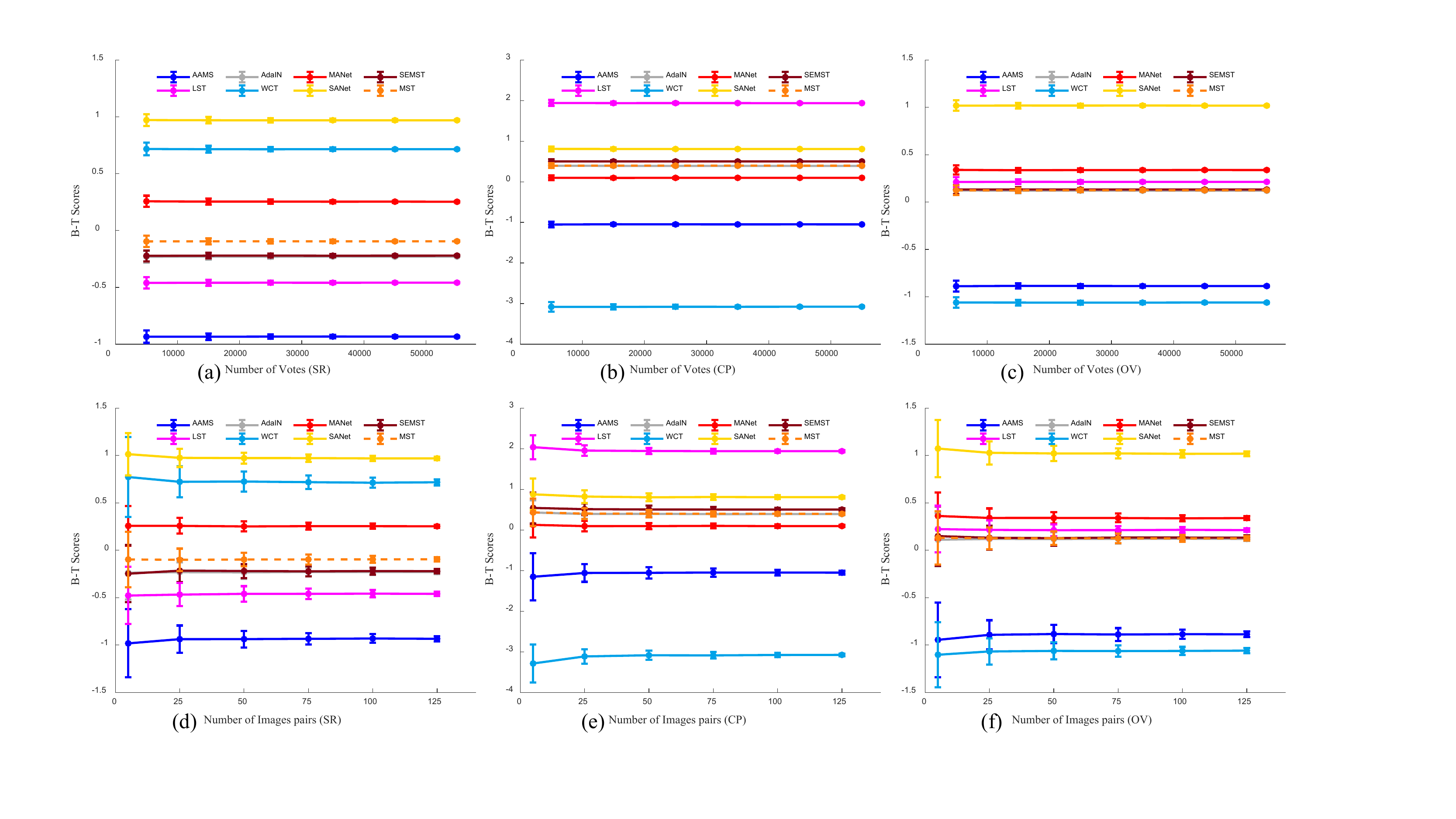}
    \caption{Convergence analysis on the number of votes and image pairs at each subjective evaluation.}
    \label{fig10}
\end{figure}

\textbf{Number of votes:} We randomly sample $\lambda$ ($\lambda$ = 5000, 15000, 25000\ldots, 55000) votes from a total of 63,000 voting results, and calculate the B-T scores for each AST algorithm. To avoid the possible bias, we repeat this process 1000 times with different samples of votes. Fig. \ref{fig10} (a)-(c) show the mean and standard deviation of B-T scores for each sample on three subjective evaluations. It is observed that as the number of votes grows, the B-T scores tend to be stable, which demonstrates that the number of votes is sufficiently large for performance evaluation.

\textbf{Number of images pairs:} Similar to the above convergence analysis, we randomly sample $\mu$ ($\mu$ = 5, 25, 50, 75, 100, 125) content-style image pairs from our dataset and then plot the means and the standard deviations in Fig. \ref{fig10} (d)-(f). Obviously, as the number of images pairs grows, standard deviation of B-T Scores decreases, which indicates that the B-T scores obtained from the subjective study are stable.

Overall, the above two kinds of convergence analysis demonstrate the reliability of the subjective rating scores. 
\begin{figure*}[!t]
    \centering
    \includegraphics[width=6.4in]{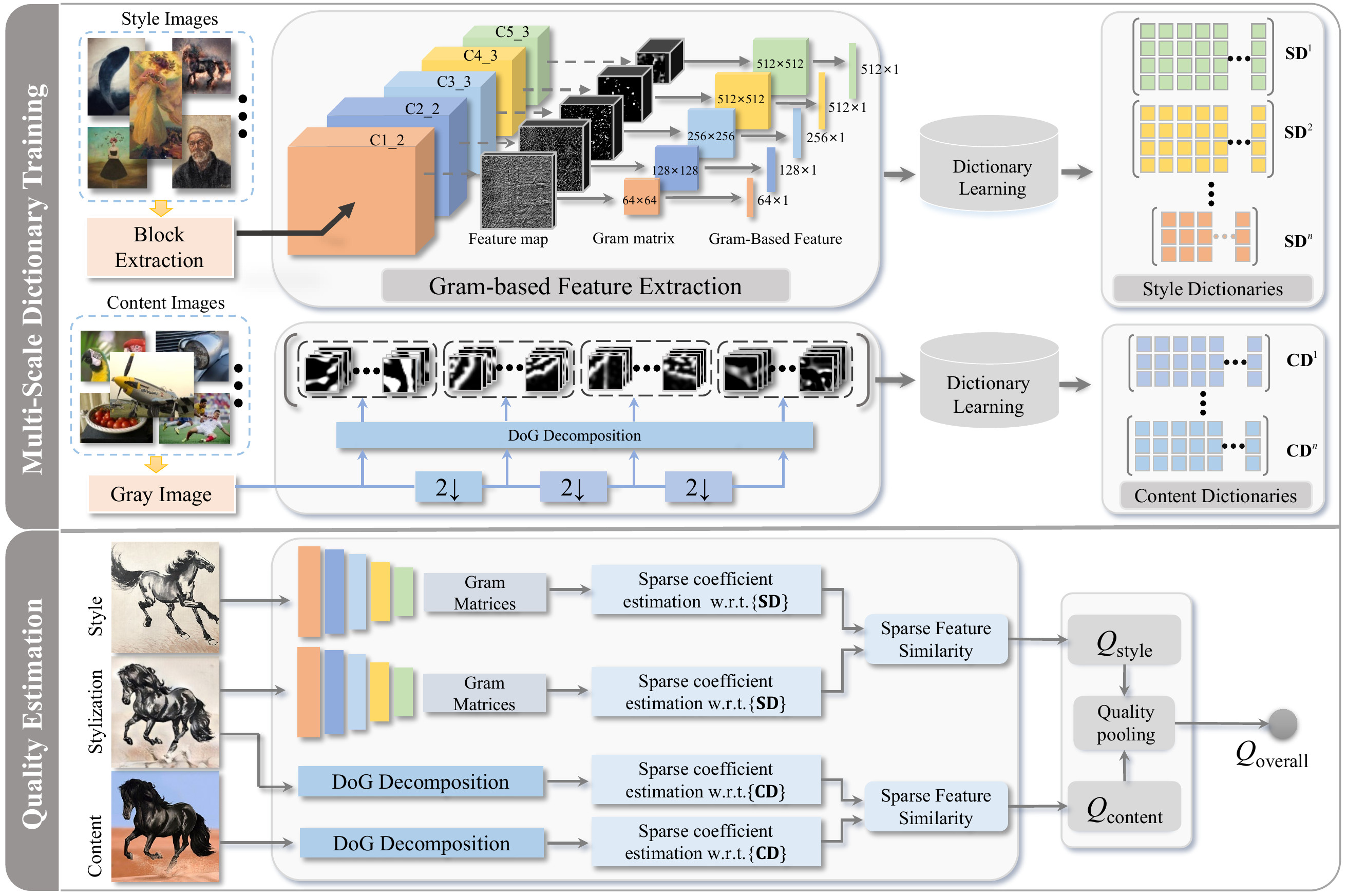}
    \caption{Framework for the proposed SRQE method.}
    \label{fig11}
\end{figure*}
\section{OBJECTIVE QUALITY EVALUATION}
In this paper, we propose a new sparse representation-based image quality evaluation metric (SRQE) for AST-IQA, as shown in Fig. \ref{fig11}. The process is composed of two phases: multi-scale dictionary training and quality estimation. In the training phase, the multi-scale style and content dictionaries, learnt from the training databases via sparse representation, are utilized to build style representation for style images and capture inherent structures for the content images, respectively. In the quality estimation phase, the quality of SR (or CP) is obtained by estimating the sparse feature similarity between the stylized image and the style (or content) image. Finally, the OV quality is acquired by combining the SR quality and CP quality together. In what follows, we elaborate on each step of the proposed method.

\subsection{Style Feature Extraction}
\begin{figure}[!t]
    \centering
    \includegraphics[width=3.3in]{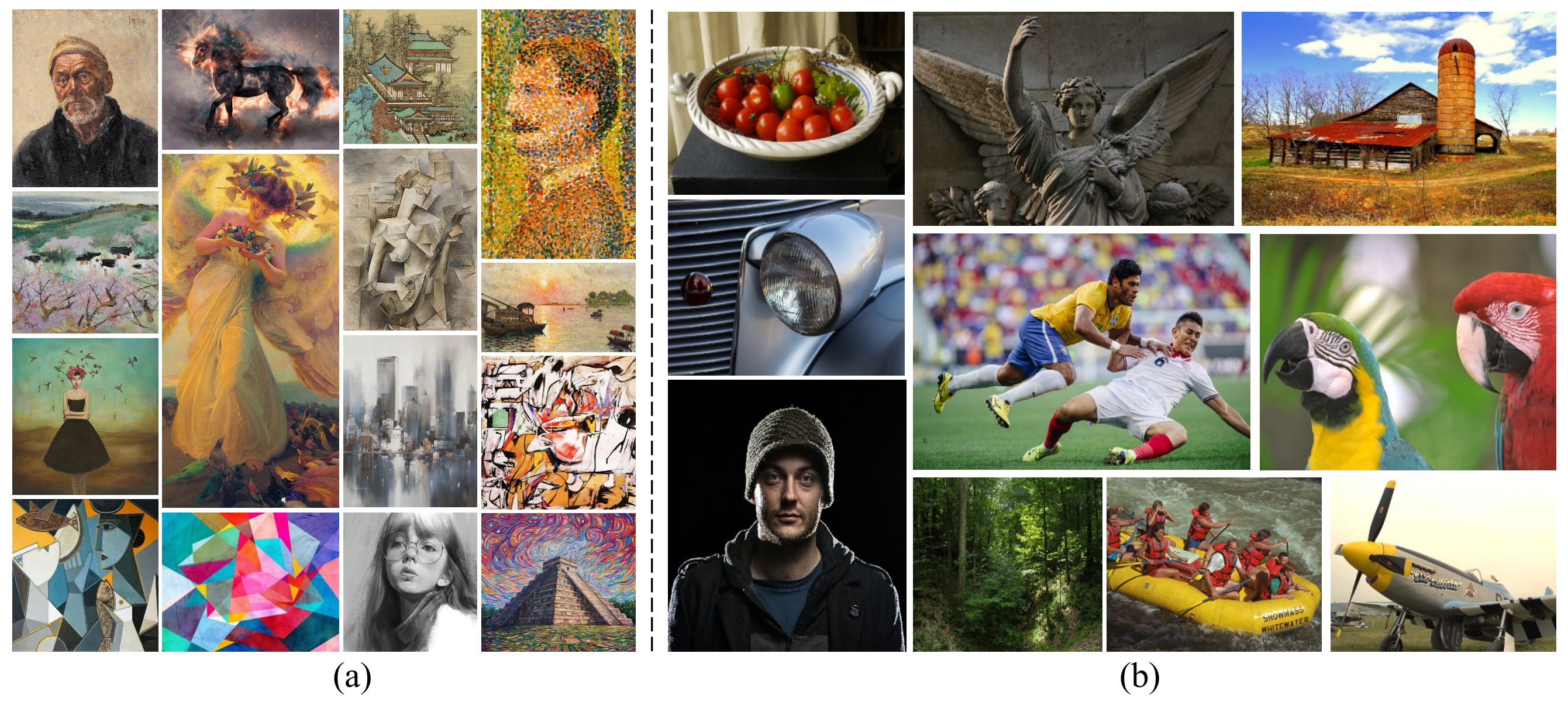}
    \caption{Some of the images of the training databases used in the paper. (a) Style training database. (b) Content training database.}
    \label{fig12}
\end{figure}

\emph{1) Selection of Training Database:} As show in Fig. \ref{fig12} (a), we re-collected 100 new style images, covering a wide variety of categories, as training images for dictionary learning. Note that there is no overlap between these images and the above collected style images (in the Section II-A) to ensure the complete independence of the training and test data. In addition, to avoid overfitting, we augment the training database by evenly partitioning the whole images. The impacts of the number of image blocks and training database will be discussed in the following Section IV-B. 

\emph{2) Gram-Based Feature:} The work in \cite{ref7} demonstrated that the correlations between convolution responses at the same layer (i.e., Gram matrices) yielded effective texture synthesis and can effectively grasp the image style. Additionally, sparse representation technique shows great prospects in image visual style analysis \cite{ref45}. Inspired by this, we construct a style perception model based on sparse representation, while incorporating high-level perceptual information (Gram matrices) extracted from deep neural network.

In this paper, we resort to compute gram matrices from style images using a pre-trained VGG network of the state-of-the-art full-reference IQA model (DISTS) \cite{ref46} which has superior performance in evaluating texture similarity. Assumed that the feature map of a sample style image $\emph{I}_s$ at layer \emph{l} of DISTS \cite{ref46} is denoted as ${{\mathbf{F} }^{l}}({{I}_{s}})\in {{\mathbb{R}}^{C\times H\times W}}$, where \emph{C} is the number of channels, and \emph{H} and \emph{W} represent the height and width of the feature map, the Gram-based representation is computed from the feature map ${{\mathbf{F} }^{l}}({{I}_{s}}{)}'\in {{\mathbb{R}}^{C\times (HW)}}$ aggregated from the ${{\mathbf{F} }^{l}}({{I}_{s}})\in {{\mathbb{R}}^{C\times H\times W}}$:
\begin{equation}
\mathbf{G} \left( {{\mathbf{F} }^{l}}({{I}_{s}}{)}' \right)=\left[ {{\mathbf{F} }^{l}}({{I}_{s}}{)}' \right]{{\left[ {{\mathbf{F} }^{l}}({{I}_{s}}{)}' \right]}^{T}}
\end{equation}

In the implementation, we use the first to fifth VGG network layers of DISTS \cite{ref46} to produce a set of Gram-based representations at different layers, namely $\text{ }\!\!\{\!\!\text{ }{{\mathbf{G}}^{l}}\in {{\mathbb{R}}^{C\times C}},\text{ }l=1,2,\ldots ,L\text{ }\!\!\}\!\!\text{ }$, where \emph{L}=5 denotes the highest layer and $C\in \left\{ 64,\text{ }128,\text{ }256,\text{ }512,\text{ }512 \right\}$ correspond to the numbers of feature maps at each layer.

After the above processing, each Gram-based representation then generates a Gram-based style feature vector ${{\mathbf{g} }^{l}}\in {{\mathbb{R}}^{C\times 1}}$ through averaging each row, described as:
\begin{equation}
\label{deqn_ex1a}
{{\mathbf{g}}}^{l}={\mathbf{G}^{l}}\cdot {\mathbf{x}^{l}}.
\end{equation}
\begin{equation}
\label{deqn_ex1a}
{\mathbf{x}^{l}}={{[x_{1}^{l},x_{2}^{l},\ldots ,x_{_{C}}^{l}]}^{T}}.
\end{equation}
where $x_{1}^{l}=\ldots =x_{C}^{l}=1/C$

Finally, numerous style feature vectors at the same layer extracted from different style images are used to form a style matrix $\mathbf{SM}$. As a result, we can obtain five different style matrices (corresponding to five layers) $\mathbf{S} {{\mathbf{M} }^{l}}=[{{\{{{\mathbf{g} }^{l}}\}}_{1}},{{\{{{\mathbf{g} }^{l}}\}}_{2}},\ldots ,{{\{{{\mathbf{g} }^{l}}\}}_{{{N}^{l}}}}]\in {{\mathbb{R}}^{C\times {{N}^{l}}}}$. All style matrices will be used for the subsequent style dictionary learning.

\emph{3) Multi-Scale Style Dictionary Learning:} Using the above style matrices $\mathbf{S} {{\mathbf{M} }^{l}}=[{{\{{{\mathbf{g} }^{l}}\}}_{1}},{{\{{{\mathbf{g} }^{l}}\}}_{2}},\ldots ,{{\{{{\mathbf{g} }^{l}}\}}_{{{N}^{l}}}}]\in {{\mathbb{R}}^{C\times {{N}^{l}}}}$ as input, we learn multi-scale style dictionary $\mathbf{SD} ^{l}$ by seeking a sparse representation for each style feature vector $\mathbf{g}^{l}$ under specific sparsity constraint $\tau$. Each style sub-dictionary $\mathbf{SD} =[\mathbf{sd}_{1},\mathbf{sd}_{2},\ldots ,\mathbf{sd}_{U}]\in {{\mathbb{R}}^{C\times U}}$ contains \emph{U} basic elements. 
Formally, the process of multi-scale style dictionary learning can be formulated as: 
\begin{equation}
\begin{aligned}
\label{deqn_ex1a}
\left\langle {{\mathbf{SD} }^{l}},\text{ }{{{\hat{\bm{\alpha }}}}_{i}} \right\rangle =\underset{{}}{\mathop{\arg \min }}\,\sum\limits_{i=1}^{N}{\left\| {{\{\mathsf{}{{\mathbf{g} }^{l}}\}}_{i}}-{{\mathbf{SD} }^{l}}{\bm{{\alpha }}_{i}} \right\|_{2}^{2}}\\\text{  s}\text{.t}\text{. }\forall i,\text{ }{{\left\| {\bm{{\alpha }}_{i}} \right\|}_{0}}\le \tau
\end{aligned}
\end{equation}
where ${{\left\| \cdot  \right\|}_{2}}$ is the $\emph{l}_2$-norm operator, ${{\left\| \cdot  \right\|}_{0}}$ denotes the $\emph{l}_0$-norm that counts the number of non-zero elements in a vector, and $\bm{\alpha}_{i}$ is the sparse coefficient vector of ${{\{{{\mathsf{\mathbf{g}}}^{l}}\}}_{i}}$. Typically, both ${{\mathbf{SD}}^{l}}$ and $\bm{\alpha}_i$ are unknown in this stage. We resort to the online dictionary learning (ODL) algorithm implemented in the SPArse Modeling Software \cite{ref47} to solve this NP-hard problem. Details of dictionary learning can refer to \cite{ref47}.

\subsection{Content Feature Extraction}
\emph{1) Selection of Training Database:} Since the essence of the proposed content evaluation model is to restore the structure information of the source content images and stylized images based on dictionary learning, we only select natural images to construct the content dictionary. Refer to \cite{ref48}, we randomly select ten natural images from the TID 2013 \cite{ref49} database and NPRgeneral \cite{ref31}, which have different scenes in the images, as shown in Fig. \ref{fig12} (b).

\begin{figure}[!t]
    \centering
    \includegraphics[width=3.3in]{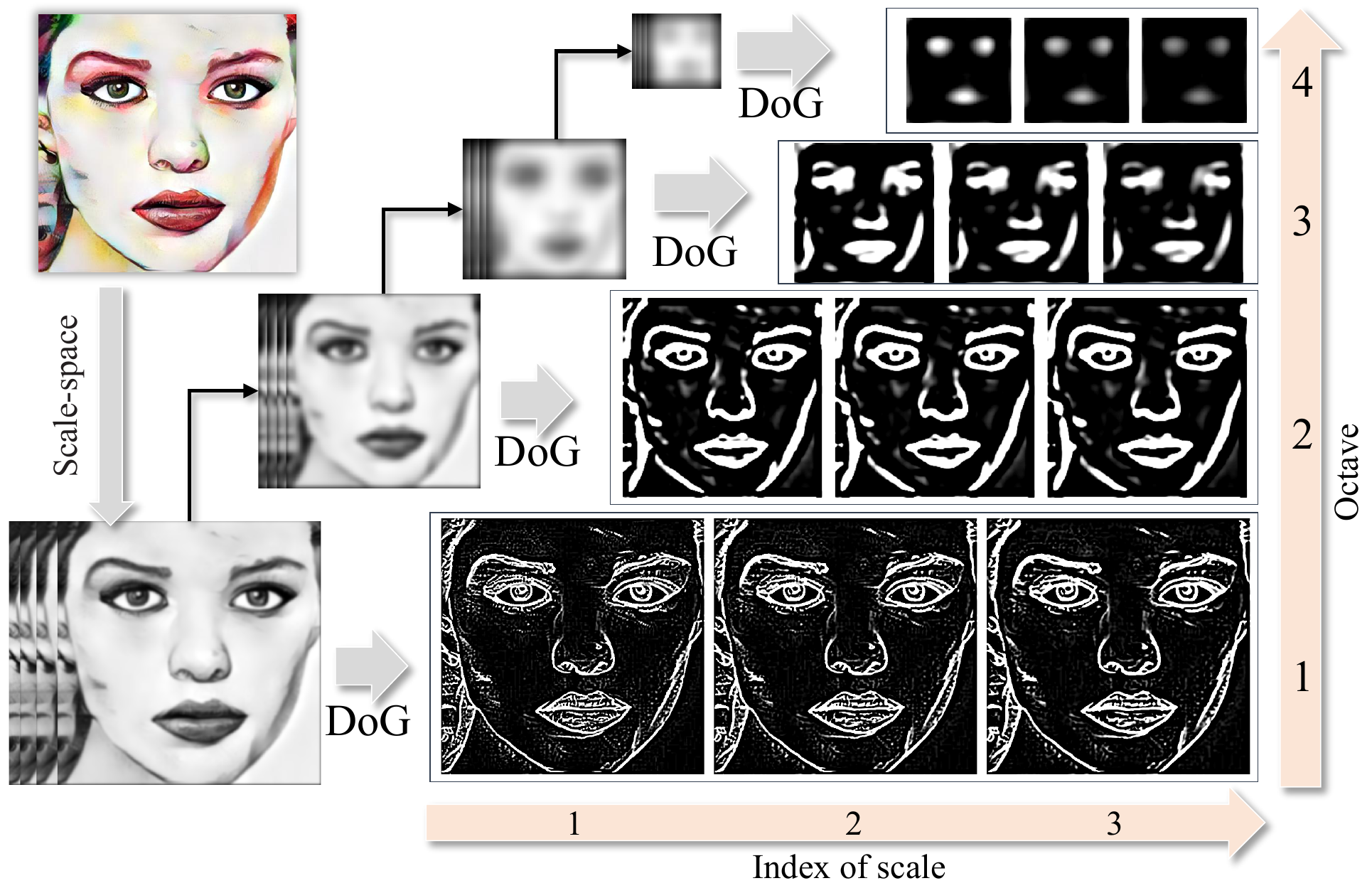}
    \caption{An example of DoG multi-scale space.}
    \label{fig13}
\end{figure}

\emph{2) DoG Response Feature:} As known, human visual perception is highly sensitive to the edge information, the major objects in the painting emphasized by the artists often contain distinct edges in most cases \cite{ref39}. Intuitively speaking, the outline of the main objects largely reflects the content information of artworks. Inspired by this, the edge information, as the significant component in painting content, needs to be deeply investigated for evaluating the CP quality. Furthermore, an outstanding painting will be appreciated by humans in local details and global perception. Thus, it is necessary to utilize multi-scale strategy to better describe the content of the painting from coarse to fine level of detail\cite{ref50}. As shown in Fig. \ref{fig13}, the multi-scale Difference of Gaussian (DoG) is applied to represent the content feature \cite{ref51}, which can properly simulate the receptive field of retinal cells.
First, the DoG signals, $DoG(x,y)$, at different scales can be computed by:
\begin{equation}
DoG(x,y)=\left| {{R}_{\sigma ,k\sigma }}(x,y)\otimes I(x,y) \right|.
\end{equation}
where $I(x,y)$ denotes the pixel location $(x,y)$ of the input image, the symbol $\otimes$ denotes the convolution operation, and ${{R}_{\sigma ,k\sigma }}(x,y)$ is defined as the difference between two Gaussian kernel with nearby scales $\sigma$ and $k\sigma$:
\begin{equation}
\begin{aligned}
{{R}_{\sigma ,k\sigma }}(x,y)=\frac{1}{2\pi {{\sigma }^{2}}}\exp (-\frac{{{x}^{2}}+{{y}^{2}}}{2{{\sigma }^{2}}})\\-\frac{1}{2\pi {{k}^{2}}{{\sigma }^{2}}}\exp (-\frac{{{x}^{2}}+{{y}^{2}}}{2{{k}^{2}}{{\sigma }^{2}}})
\end{aligned}
\end{equation}
where $\sigma$ and $k$ are used to control the scales of DoG. Refer to \cite{ref48}, we set $k$ = 1.6, and $\sigma \in \left\{ 0,\text{ }1,\text{ }1.6,\text{ }2.56,\text{ }4.096 \right\}$ in the experiment. Here, $\sigma$ = 0 denotes the original scale.

Once the DoG signals at the current octave are computed, the last scale-space image was selected as the new input and was down-sampled by a factor of two to repeat the above process, thereby producing a set of DoG signals with a variety of octaves and scales, namely $\left\{ Do{{G}^{z,o}}(x,y) \right\}$, where $z\in \{1,\ldots ,Z\}$ denotes the $z$-th scale, and $o\in \{1,\ldots ,O\}$ denotes $o$-th octave.

After the above processing, each DoG signal is partitioned into numerous patches with size of $8\times8$, subtracted by the mean value. In the implementation, 1000 overlapped patches having rich details and structures are selected as training samples. Then, these patches are vectorized into column vectors to form a content matrix $\mathbf{CM}$, $\mathrm{\mathbf{CM} }=[{{\mathrm{\mathbf{y} }}_{1}},\ldots ,{{\mathrm{\mathbf{y} }}_{k}}]\in {{\mathbb{R}}^{T\times K}}$, based on which the subsequent overcomplete content dictionary is learned. Each patch ${{\mathrm{\mathbf{y} }}_{k}}\in {{\mathbb{R}}^{T\times 1}}$ contains $T$ pixels and $k=1,\ldots ,K$. Here, $K$ = 1000.

\emph{3) Multi-Scale Content Dictionary Learning:} Similar to the above multi-scale style dictionary learning, the multi-scale content dictionary $\mathbf{CD}^{z,o}$ can be learned from multi-scale content matrices $\mathbf{CM}^{z,o}$. Each content sub-dictionary $\mathbf{CD} =[\mathbf{cd}_{1},\mathbf{cd}_{2},\ldots ,\mathbf{cd}_{U}]\in {{\mathbb{R}}^{C\times U}}$ contains $V$ basic elements. In the experiment, we set $V$ = 256. Similarly, the process of multi-scale content dictionary learning can be formulated as:
\begin{equation}
\begin{aligned}
\left\langle {\mathbf{CD} }^{z,o},\text{ }{{{\hat{\bm{\beta }}}}_{i}} \right\rangle =\underset{{}}{\mathop{\arg \min }}\,\sum\limits_{i=1}^{K}{\left\| {{\mathrm{\mathbf{y} }}_{i}}-{{\mathbf{CD} }^{z,o}}{\bm{\beta }_{i}} \right\|_{2}^{2}}\\\text{  s}\text{.t}\text{. }\forall i,\text{ }{{\left\| {\bm{\beta }_{i}} \right\|}_{0}}\le \tau
\end{aligned}
\end{equation}
where $\bm{\beta}_i$ is the sparse coefficient vector of $\mathbf{y}_i$. Note that we also apply the ODL algorithm to solve Eq. (10)

\subsection{Feature Similarity Measurement}
Through the above efforts, we obtain two types of over-complete multi-scale dictionaries, containing $U$ and $V$ basic atoms as the column vectors in ${{\mathbf{SD}}^{l}}$ and ${{\mathbf{CD}}^{z,o}}$, respectively. Thus, each style feature vector ${{{\mathbf{g}}}^{l}}$ (or content patch $\mathbf{y}_k$) can be sparsely represented as a linear combination of basic atoms contained in ${{\mathbf{SD}}^{l}}$ (or ${{\mathbf{CD}}^{z,o}}$).

\emph{1) Style Sparse Coefficients Estimation:} Given the testing stylized image $I_t$ and style image $I_s$, we can obtain two Gram-based representations using the DISTS network, denoted as $\mathbf{G}_{t}^{l}\in {{\mathbb{R}}^{C\times C}}$ and $\mathbf{G}_{s}^{l}\in {{\mathbb{R}}^{C\times C}}$. Then, the style sparse coefficient vectors can be estimated by a weighted linear combination of previously learnt dictionary elements, i.e.,
\begin{equation}
{{\mathbf{s}}^{l}}=\mathbf{G}_{s}^{l}\times {{(\mathbf{SD}^{l})}^{+}}
\end{equation}
\begin{equation}
{{\mathbf{ts}}^{l}}=\mathbf{G}_{t}^{l}\times {{(\mathbf{SD}^{l})}^{+}}
\end{equation}
where ${(\mathbf{SD}^{l})}^{+}$ denotes the generalized inverse matrices of ${(\mathbf{SD}^{l})}$.

\emph{2) Content Sparse Coefficients Estimation:} For the testing stylized image $I_t$ and the source content image $I_c$, after the same processing steps as in the training phase, we can obtain patch vectors $\mathrm{\mathbf{y}}_{c}^{z,o}$ from $I_c$ and its corresponding patch vectors $\mathrm{y}_{t}^{z,o}$ from $I_t$. Similarly, the content sparse coefficient \textbf{c} and \textbf{tc} can be computed by:
\begin{equation}
{{\mathbf{c}}^{z,o}}=\mathbf{y}_{c}^{z,o}\times {{(\mathbf{CD}^{z,o})}^{+}}
\end{equation}
\begin{equation}
{{\mathbf{tc}}^{z,o}}=\mathbf{y}_{t}^{z,o}\times {{(\mathbf{CD}^{z,o})}^{+}}
\end{equation}
where ${(\mathbf{CD}^{l})}^{+}$ denote the generalized inverse matrices of ${(\mathbf{CD}^{l})}$.

\emph{3) Sparse Feature Similarity Measure:} From the above estimation phase, we generate the sparse coefficients ${{\mathbf{s}}^{l}},{{\mathbf{c}}^{z,o}},\mathbf{ts}_{{}}^{l},\mathbf{tc}_{{}}^{z,o}$ on style, content and stylized images. Considering that these sparse coefficients are represented as a linear combination of basis vectors, meaning that the similarity between the style feature vectors or content patches can be directly measured using their sparse coefficient vectors. Thus, the style and content similarities are respectively defined as:
\begin{equation}
SS^{l}\left[ \mathbf{G} _{s}^{l},\mathbf{G} _{t}^{l} \right]=\frac{2\left\langle {{\mathbf{s}}^{l}},\mathbf{ts}_{{}}^{l} \right\rangle +\eta }{{{\left\| {{\mathrm{\mathbf{s} }}^{l}} \right\|}_{2}}\cdot {{\left\| \mathbf{ts}_{{}}^{l} \right\|}_{2}}+\eta }
\end{equation}
\begin{equation}
CS^{z,o}\left[ \mathbf{y}_{c}^{z,o},\mathbf{y}_{t}^{z,o} \right]=\frac{2\left\langle {{\mathbf{c}}^{z,o}},\mathbf{tc}_{{}}^{z,o} \right\rangle +\eta }{{{\left\| {{\mathbf{c}}^{z,o}} \right\|}_{2}}\cdot {{\left\| \mathbf{tc}_{{}}^{z,o} \right\|}_{2}}+\eta }
\end{equation}
where $\left\langle \cdot  \right\rangle$ calculates the inner product, $\eta$ is a constant with a small value added to prevent the denominator to be zero. The $SS$ measures the style similarity between the style and the stylized image, and $CS$ measures the content similarity between the content and the stylized image.
\subsection{Final Quality Pooling}
To measure the final quality between a stylized image and its corresponding content and style images, we need to pool the above spare feature similarities into a single score. In our pooling strategy, we first pool the style and content sparse feature similarities into SR and CP scores across all scales or octaves, and then combine the scores to measure the OV quality score. First, the SR quality score $Q_{\mathrm{style}}$ is defined as:
\begin{equation}
{{Q}_{\mathrm{style}}}=\prod\limits_{l=1}^{L}{{{\left( SS \right)}^{l}}}
\end{equation}

Then, the CP quality score $Q_{\mathrm{content}}$ is defined as:
\begin{equation}
{{Q}_{\mathrm{content}}}=\frac{1}{{{Z}^{2}}}\prod\limits_{o=1}^{O}{\left( \sum\limits_{z=1}^{Z}{\left( CS \right)_{{}}^{z,o}} \right)}
\end{equation}
where $Z$ and $O$ denote the number of scales and octaves.

Finally, the OV quality $Q_{\mathrm{overall}}$ is calculated by combining $Q_{\mathrm{style}}$ and $Q_{\mathrm{content}}$ into a score:
\begin{equation}
\label{Qoverall}
{{Q}_{\mathrm{overall}}}={{({{Q}_{\mathrm{content}}})}^{{{\omega }_{1}}}}\cdot {{({{Q}_{\mathrm{style}}})}^{{{\omega }_{2}}}}
\end{equation}
where the parameters $\omega_1$ and $\omega_2$ are used to adjust the relative importance of the two portions. In this paper, we set $\omega_1$ = 0.4 and $\omega_2$ = 0.6 based on the performance analyses in Section IV-C. Of course, there is a large room to manipulate the importance weights for better quality prediction. A more meaningful practice may be to explore the proper combination of $Q_{\mathrm{style}}$ and $Q_{\mathrm{content}}$ that best fits human subjective study.

\section{EXPERIMENTAL RESULTS}
\subsection{Evaluation Criteria}
Similar to \cite{ref52,ref53,ref54}, four criteria are adapted to measure the performance of different methods: the Spearman Rank order Correlation Coefficient (SRCC), Kendall Rank-order Correlation Coefficient (KRCC), Pearson Linear Correlation Coefficient (PLCC), and Hit Rate (HITR). Specifically, the SRCC and KRCC measure the prediction monotonicity. PLCC is utilized to evaluate the prediction linearity after fitting a five-parameter logistic function:
\begin{equation}
f(x)={{\kappa }_{1}}\left( \frac{1}{2}-\frac{1}{1+\exp ({{\kappa }_{2}}(x\text{-}{{\kappa }_{\text{3}}}))} \right)\text{+}{{\kappa }_{\text{4}}}\cdot x\text{+}{{\kappa }_{\text{5}}}
\end{equation}
where $x$ and $f(x)$ represent the objective and mapped scores respectively, and $\{{{\kappa }_{i}}|i=1,2,\ldots ,5\}$ are the five parameters to be fitted.
Additionally, HITR can measure the classification accuracy \cite{ref52}, defined as:
\begin{equation}
HITR={{R}_{i}}/{{R}_{n}}\
\end{equation}
where $R_{i}$ denotes the number of correct judgments in PC, and $R_{n}$ is the total number of PC. Considering that the subjective experiments are based on PC within the same group of images, so that the ground truth (B-T scores) are only meaningful within the same group. Therefore, these criteria are computed respectively for each group from the same source image. Then, the average value of all 150 groups is reported as the final performance score. A superior metric should have higher criteria values (with a maximum of 1).
\begin{figure}[!t]
    \centering
    \includegraphics[width=3.3in]{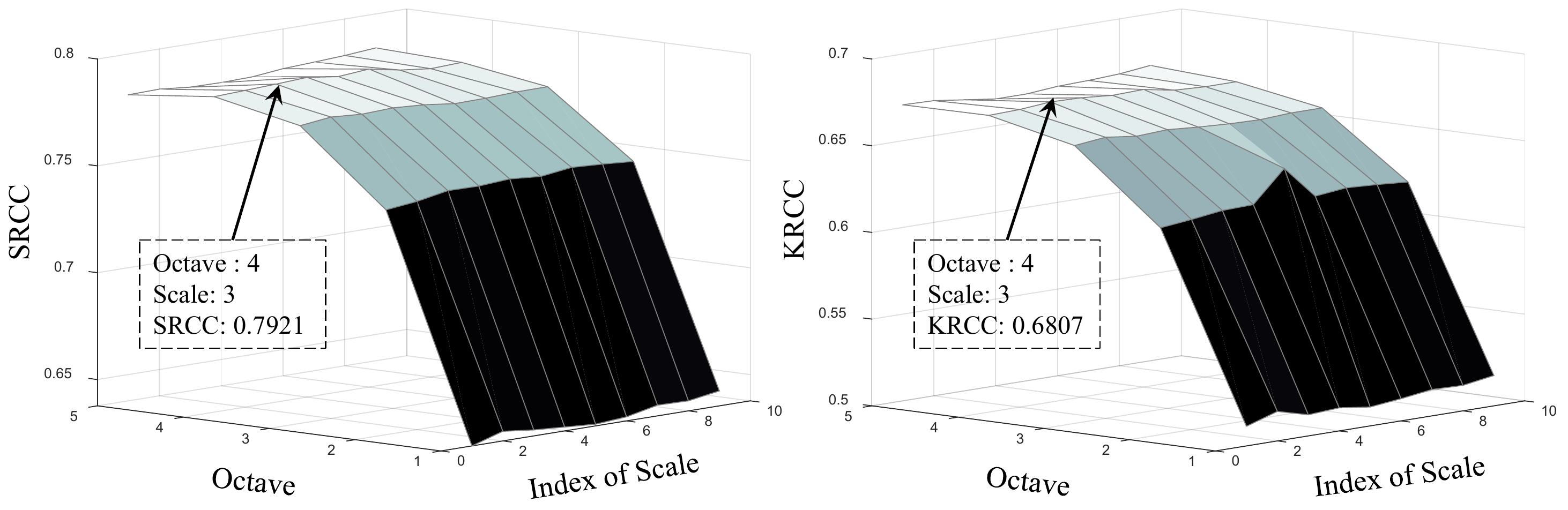}
    \caption{Impacts of various octave and scale combinations on the performance of the $Q_{\mathrm{content}}$ on CP evaluation.}
    \label{fig14}
\end{figure}
\begin{figure}[!t]
    \centering
    \includegraphics[width=3.3in]{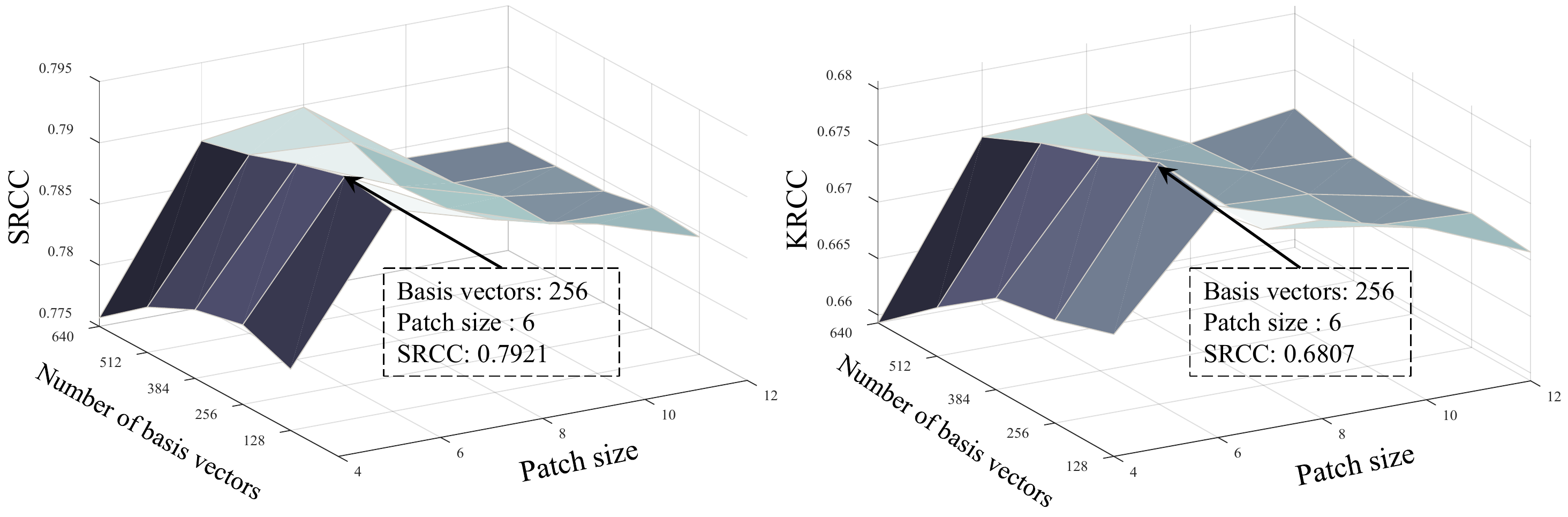}
    \caption{Impacts of various patch sizes and numbers of basis vectors combinations on the performance of the $Q_{\mathrm{content}}$ on CP evaluation.}
    \label{fig15}
\end{figure}
\begin{figure}[!t]
    \centering
    \includegraphics[width=3.3in]{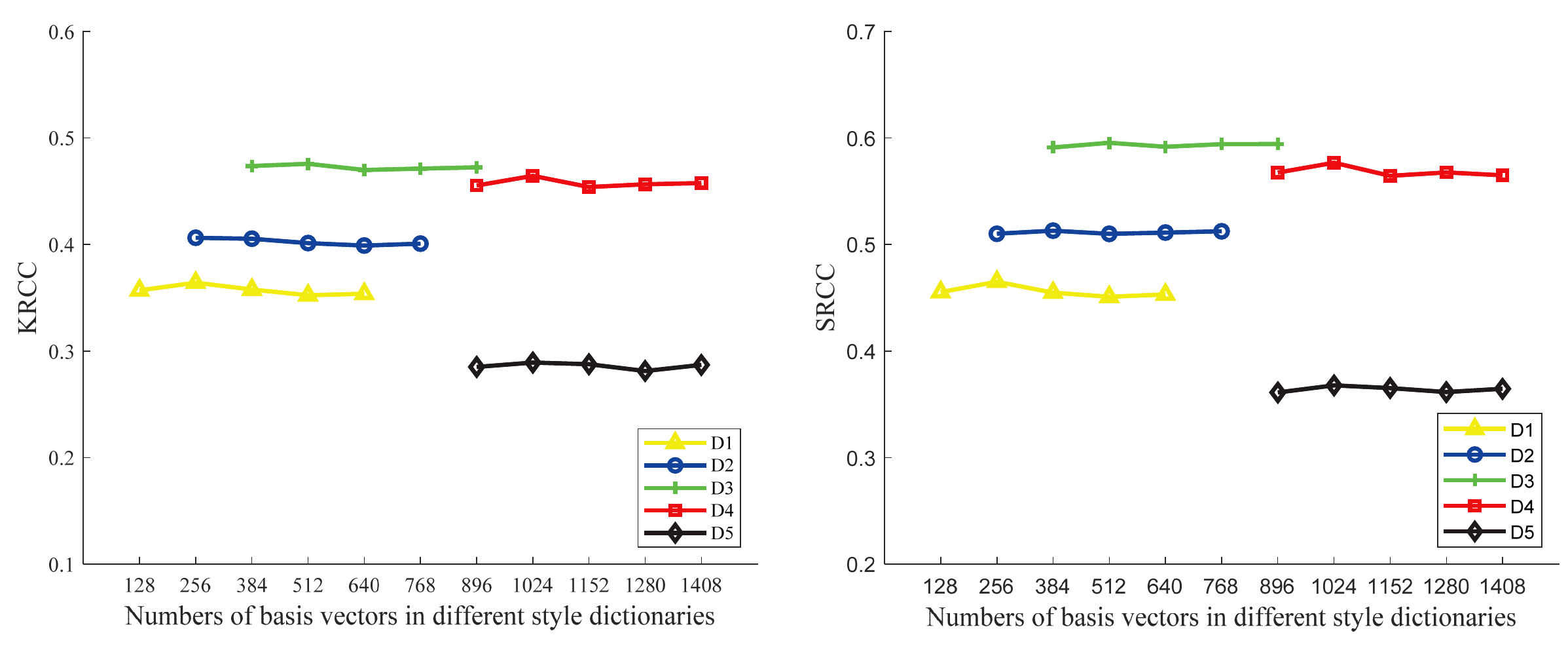}
    \caption{Impacts of numbers of basis vectors on the performance of the $Q_{\mathrm{style}}$ on SR evaluation.}
    \label{fig16}
\end{figure}
\begin{table*}[t]
\renewcommand{\arraystretch}{1.2} 
\caption{Performance comparison of CP and SR for the proposed method with different training databases.}
\label{tab2}
\begin{center}
\resizebox{2\columnwidth}{!}{
\begin{tabular}{c|c|c|c|c|cccc}
\hline\hline
Evaluation          & Dictionary & Database            & Training images & Training samples for each sub-dictionary & SRCC   & KRCC   & HITR   & PLCC   \\ \hline
\multirow{4}{*}{CP} & Dict. I    & TID 2013            & 10              & 1000 (patches)                           & 0.7906 & 0.6773 & 0.8376 & 0.8637 \\ 
                    & Dict. II    & NPRgeneral          & 10              & 1000 (patches)                           & 0.7903 & 0.6783 & 0.8383 & 0.8637 \\  
                    & Dict. III    & TID 2013+NPRgeneral & 40              & 4000 (patches)                           & 0.7900 & 0.6774 & 0.8379 & 0.8638 \\ 
                    & Dict. IV    & Proposed            & 10              & 1000 (patches)                           & 0.7921 & 0.6807 & 0.8393 & 0.8635 \\ \hline
\multirow{3}{*}{SR} & Dict. I    & TAD66K-art          & 100             & 400, 400, 900, 1600, 1600 (vectors)      & 0.6034 & 0.4827 & 0.7386 & 0.6236 \\ 
                    & Dict. II    & TAD66K-art          & 400             & 1600, 1600, 3600, 6400, 6400 (vectors)   & 0.6028 & 0.4816 & 0.7378 & 0.6227 \\ 
                    & Dict. III    & Proposed            & 100             & 400, 400, 900, 1600, 1600  (vectors)     & 0.6062 & 0.4886 & 0.7412 & 0.6278 \\ \hline\hline
\end{tabular}
}
\end{center}
\end{table*}
\begin{table*}[t]
\renewcommand{\arraystretch}{1.2} 
\caption{The impacts of different pooling strategy on the performance of the proposed method.}
\label{tab3}
\begin{center}
\resizebox{2\columnwidth}{!}{
\begin{tabular}{c|ccccc|ccccc|ccccc}
\hline\hline
\multirow{3}{*}{\begin{tabular}[c]{@{}c@{}}Pooling\\ strategy\end{tabular}} & \multicolumn{5}{c|}{Multiplication (c, d)=(1,0)}                                                                                              & \multicolumn{5}{c|}{Summation (c, d)=(0,1)}                                                                                                   & \multicolumn{5}{c}{Combination (w1, w2, w3, w4)=( 0.4, 0.6, 0.4, 0.6)}                                                                       \\ \cline{2-16} 
                                                                            & \multicolumn{5}{c|}{(w1, w2)}                                                                                                                 & \multicolumn{5}{c|}{(w3, w4)}                                                                                                                 & \multicolumn{5}{c}{(c, d)}                                                                                                                   \\ \cline{2-16} 
                                                                            & \multicolumn{1}{c|}{(0.5,0.5)} & \multicolumn{1}{c|}{(0.6,0.4)} & \multicolumn{1}{c|}{(0.8,0.2)} & \multicolumn{1}{c|}{(0.4,0.6)} & (0.2,0.8) & \multicolumn{1}{c|}{(0.5,0.5)} & \multicolumn{1}{c|}{(0.6,0.4)} & \multicolumn{1}{c|}{(0.8,0.2)} & \multicolumn{1}{c|}{(0.4,0.6)} & (0.2,0.8) & \multicolumn{1}{c|}{(0.5,0.5)} & \multicolumn{1}{c|}{(0.6,0.4)} & \multicolumn{1}{c|}{(0.8,0.2)} & \multicolumn{1}{c|}{(0.4,0.6)} & (0.2,0.8) \\ \hline
SRCC                                                                        & \multicolumn{1}{c|}{0.5980}    & \multicolumn{1}{c|}{0.5666}    & \multicolumn{1}{c|}{0.5053}    & \multicolumn{1}{c|}{\textbf{0.6077}}    & 0.4495    & \multicolumn{1}{c|}{0.5637}    & \multicolumn{1}{c|}{0.5397}    & \multicolumn{1}{c|}{0.4892}    & \multicolumn{1}{c|}{0.5990}    & 0.4894    & \multicolumn{1}{c|}{0.5985}    & \multicolumn{1}{c|}{0.5997}    & \multicolumn{1}{c|}{0.5984}    & \multicolumn{1}{c|}{0.5968}    & 0.6046    \\ 
KRCC                                                                        & \multicolumn{1}{c|}{0.4746}    & \multicolumn{1}{c|}{0.4493}    & \multicolumn{1}{c|}{0.3920}    & \multicolumn{1}{c|}{\textbf{0.4855}}    & 0.3563    & \multicolumn{1}{c|}{0.4502}    & \multicolumn{1}{c|}{0.4254}    & \multicolumn{1}{c|}{0.3792}    & \multicolumn{1}{c|}{0.4760}    & 0.3989    & \multicolumn{1}{c|}{0.4779}    & \multicolumn{1}{c|}{0.4793}    & \multicolumn{1}{c|}{0.4784}    & \multicolumn{1}{c|}{0.4754}    & 0.4836    \\ 
HITR                                                                        & \multicolumn{1}{c|}{0.7362}    & \multicolumn{1}{c|}{0.7243}    & \multicolumn{1}{c|}{0.6964}    & \multicolumn{1}{c|}{\textbf{0.7410}}    & 0.6776    & \multicolumn{1}{c|}{0.7248}    & \multicolumn{1}{c|}{0.7124}    & \multicolumn{1}{c|}{0.6900}    & \multicolumn{1}{c|}{0.7367}    & 0.6938    & \multicolumn{1}{c|}{0.7374}    & \multicolumn{1}{c|}{0.7381}    & \multicolumn{1}{c|}{0.7376}    & \multicolumn{1}{c|}{0.7360}    & 0.7400    \\ 
PLCC                                                                        & \multicolumn{1}{c|}{0.6668}    & \multicolumn{1}{c|}{0.6500}    & \multicolumn{1}{c|}{0.5792}    & \multicolumn{1}{c|}{\textbf{0.6510}}    & 0.4704    & \multicolumn{1}{c|}{0.6445}    & \multicolumn{1}{c|}{0.6168}    & \multicolumn{1}{c|}{0.5593}    & \multicolumn{1}{c|}{0.6561}    & 0.5277    & \multicolumn{1}{c|}{0.6581}    & \multicolumn{1}{c|}{0.6584}    & \multicolumn{1}{c|}{0.6578}    & \multicolumn{1}{c|}{0.6575}    & 0.6551    \\ \hline\hline
\end{tabular}
}
\end{center}
\end{table*}
\subsection{Parameter and Training Database Setting}
Since the fundamental of proposed quality metrics (i.e., $Q_{\mathrm{style}}$, $Q_{\mathrm{content}}$ and $Q_{\mathrm{overall}}$ ) is the dictionary operating in a multi-scale framework with several parameters, it makes sense to explore the influences of parameters and training databases on the performance evaluation. Therefore, this subsection first tunes the multi-scale dictionaries parameters, and then changes the type and number of images in the training databases. 

\emph{1) Parameters in $Q_{\mathrm{content}}$:} For multi-scale content dictionary, we first visualize the influence of different combinations of scale and octave with at most 10 scales and 5 octaves on the performance of $Q_{\mathrm{content}}$ in Fig. \ref{fig14}. It shows that the performance is greatly affected by the number of octaves while slightly affected by the number of scales. The reason is that the CP evaluation is more concerned with semantic structural changes rather than detail fidelity. As shown in the Fig. \ref{fig13}, octave adjustment causes large structural changes, while scale adjustment mainly affects small detail information. Here, we set the number of octaves $O$ = 4 and the number of scales $Z$ = 3, which can achieve the best performance. In addition, we show the performance of tuning the patch size and the number of basis vectors in Fig. \ref{fig15}. From the results, we can observe that the optimal performance is obtained by selecting patch size 6 and the number of basis vectors 256. Therefore, we set $T$ = 36 (patch size = 6) and $V$ = 256 in this work.

\emph{2) Parameters in $Q_{\mathrm{style}}$:} For multi-scale style dictionary, since the Gram-based style feature vector of each layer is fixed, we only test the effect of the number of basis vectors. As shown in Fig. \ref{fig16}, the evaluation accuracy varies relatively slight over the numbers of basis vectors, which indicates that the $Q_{\mathrm{style}}$ does not highly depend on the training configurations. In this paper, we set $U\in \{256,256,512,1024,1024\}$.

\emph{3) Training database:} In addition to the parameters in the dictionary, the selection of the training database is also worth analyzing, which can validate whether the performance is dependent on a particular training database. To this end, we train several overcomplete content and style dictionaries based on different training images respectively. The training images are selected from the TID 2013 \cite{ref49}, NPRgeneral \cite{ref31}, and TAD66K \cite{ref55}. The implementation details and performance results are listed in Table \ref{tab2}. Note that the number of training samples is maintained consistent for each content sub-dictionary, while the training samples (i.e., Gram-based style feature vectors) of style sub-dictionaries are set to a different number to match the size of the Gram matrix. The results represent that the performance with different dictionaries are quite similar. This demonstrates that the proposed method is insensitive to the selection of the training databases.

\subsection{Analysis of Pooling Methods}
As mentioned above, there is a complex relationship between OV and other quality factors (e.g., $Q_{\mathrm{style}}$ and $Q_{\mathrm{content}}$). Hence, it is meaningful to analyze the impacts of different pooling methods on the performance results. In this connection, we refer to the experimental settings \cite{ref56} making a modification to Equation (\ref{Qoverall}):
\begin{equation}
\begin{aligned}
{{Q}_{\mathrm{overall}}}=c\times ({{Q}_{\mathrm{content}}}^{{{\omega }_{1}}})\times ({{Q}_{\mathrm{style}}}^{{{\omega }_{2}}})+\\
d\times ({{{\omega }_{3}}}{{Q}_{\mathrm{content}}}+{{{\omega }_{4}}}{{Q}_{\mathrm{style}}})\
\end{aligned}
\end{equation}
where $c$ and $d$ are utilized to balance the significance of the summation and multiplication pooling items, $\omega_1$, $\omega_2$, $\omega_3$ and $\omega_4$ are set to balance the importance of different quality factors. The performance of the three pooling strategy (i.e., multiplication($c$=1, $d$=0), summation ($c$=0, $d$=1) and combination) are listed in the Table \ref{tab3}. From the results, we can find that both multiplication and summation strategies achieve the optimal performance when ($\omega_1$, $\omega_2$) and ($\omega_3$, $\omega_4$) are set as (0.4, 0.6), but the performance of multiplication is better than that of summation. Based on the optimal weight setting for the quality factors, the combination pooling strategy obtains similar results among the five ($c$, $d$) combinations. In summary, different pooling strategies generate significant impacts on the performance, in contrast, the multiplication pooling strategy with parameter ($\omega_1$=0.4, $\omega_2$=0.6) obtains the optimal results. As a consequence, we employ this multiplication pooling strategy as the final pooling method in this work.

\subsection{Performance Test on Different Quality Factors}
For the performance test on the CP, SR and OV quality evaluations, we compare the proposed SRQE with existing IQA metrics, including two categories: (1) Fourteen state-of-the-art general-purpose FR-IQA metrics: SSIM \cite{ref35}, FSIM \cite{ref26}, MS-SSIM \cite{ref57}, IW-SSIM \cite{ref25}, Peak Signal-to-Noise Ratio (PSNR), MAD \cite{ref58}, VIF \cite{ref24}, VSI \cite{ref27}, GMSD \cite{ref59}, UQI \cite{ref60}, IFC \cite{ref61}, RFSIM \cite{ref62}, DISTS \cite{ref46} and LPIPS \cite{ref63}. (2) Eight state-of-the-art general-purpose NR-IQA metrics: NIQE \cite{ref28}, TCLT \cite{ref36}, BMPRI \cite{ref64}, BLIINDS-II \cite{ref65}, BRISQUE \cite{ref37}, UNIQUE \cite{ref66}, WaDIQaM \cite{ref67}, TReS \cite{ref68}. For the traditional learning-based models, we randomly divide the each individual dataset into two non-overlapping subsets (80\% for training and 20\% for testing). Then, we resort to the support vector regression (SVR) to train the models and report the average results after training-testing process 1000 times. For the deep learning-based methods, we divide the data set into five fixed subsets with non-overlapping contents, and use four subsets for training and one subset for testing at each time to ensure that each image has been tested. For the methods that are training-free or require specific variance of human opinions, we directly report the performance results using the pre-trained model.
\begin{table*}[t]
\renewcommand{\arraystretch}{1.2} 
\caption{Performance comparison on CP evaluation. ‘*’ indicates that the method is re-trained on the AST-IQAD.}
\label{tab4}
\begin{center}
\resizebox{2\columnwidth}{!}{
\begin{tabular}{cc|cccc|cccc|cccc|cccc|cccc|cccc}
\hline\hline
\multicolumn{2}{c|}{\multirow{2}{*}{Method}}              & \multicolumn{4}{c|}{Portrait}                                                                    & \multicolumn{4}{c|}{Building}                                                                    & \multicolumn{4}{c|}{Nature}                                                                      & \multicolumn{4}{c|}{Animal}                                                                      & \multicolumn{4}{c|}{Daily Life}                                                                  & \multicolumn{4}{c}{All}                                                                         \\ \cline{3-26} 
\multicolumn{2}{c|}{}                                     & \multicolumn{1}{c|}{SRCC}   & \multicolumn{1}{c|}{KRCC}   & \multicolumn{1}{c|}{HITR}   & PLCC   & \multicolumn{1}{c|}{SRCC}   & \multicolumn{1}{c|}{KRCC}   & \multicolumn{1}{c|}{HITR}   & PLCC   & \multicolumn{1}{c|}{SRCC}   & \multicolumn{1}{c|}{KRCC}   & \multicolumn{1}{c|}{HITR}   & PLCC   & \multicolumn{1}{c|}{SRCC}   & \multicolumn{1}{c|}{KRCC}   & \multicolumn{1}{c|}{HITR}   & PLCC   & \multicolumn{1}{c|}{SRCC}   & \multicolumn{1}{c|}{KRCC}   & \multicolumn{1}{c|}{HITR}   & PLCC   & \multicolumn{1}{c|}{SRCC}   & \multicolumn{1}{c|}{KRCC}   & \multicolumn{1}{c|}{HITR}   & PLCC   \\ \hline
\multicolumn{1}{c|}{\multirow{5}{*}{NR-IQA}}      & NIQE      & \multicolumn{1}{c|}{0.0490} & \multicolumn{1}{c|}{0.0290} & \multicolumn{1}{c|}{0.5145} & 0.0258 & \multicolumn{1}{c|}{0.0565} & \multicolumn{1}{c|}{0.0623} & \multicolumn{1}{c|}{0.5310} & 0.0376 & \multicolumn{1}{c|}{0.0534} & \multicolumn{1}{c|}{0.0699} & \multicolumn{1}{c|}{0.5357} & 0.0899 & \multicolumn{1}{c|}{0.0506} & \multicolumn{1}{c|}{0.0446} & \multicolumn{1}{c|}{0.5226} & 0.0529 & \multicolumn{1}{c|}{0.0907} & \multicolumn{1}{c|}{0.0687} & \multicolumn{1}{c|}{0.5357} & 0.1158 & \multicolumn{1}{c|}{0.0601} & \multicolumn{1}{c|}{0.0549} & \multicolumn{1}{c|}{0.5281} & 0.0644 \\ 
\multicolumn{1}{c|}{}                         & TCLT      & \multicolumn{1}{c|}{0.2546} & \multicolumn{1}{c|}{0.1957} & \multicolumn{1}{c|}{0.5964} & 0.4184 & \multicolumn{1}{c|}{0.2941} & \multicolumn{1}{c|}{0.2344} & \multicolumn{1}{c|}{0.6131} & 0.3564 & \multicolumn{1}{c|}{0.3009} & \multicolumn{1}{c|}{0.2252} & \multicolumn{1}{c|}{0.3340} & 0.6119 & \multicolumn{1}{c|}{0.2398} & \multicolumn{1}{c|}{0.1800} & \multicolumn{1}{c|}{0.5881} & 0.3812 & \multicolumn{1}{c|}{0.1131} & \multicolumn{1}{c|}{0.0643} & \multicolumn{1}{c|}{0.5321} & 0.2986 & \multicolumn{1}{c|}{0.2405} & \multicolumn{1}{c|}{0.1799} & \multicolumn{1}{c|}{0.5721} & 0.3577 \\ 
\multicolumn{1}{c|}{}                         & BMPRI*    & \multicolumn{1}{c|}{0.4887} & \multicolumn{1}{c|}{0.3942} & \multicolumn{1}{c|}{0.6966} & 0.5327 & \multicolumn{1}{c|}{0.4549} & \multicolumn{1}{c|}{0.3485} & \multicolumn{1}{c|}{0.6752} & 0.5678 & \multicolumn{1}{c|}{0.3672} & \multicolumn{1}{c|}{0.2741} & \multicolumn{1}{c|}{0.6373} & 0.4255 & \multicolumn{1}{c|}{0.3927} & \multicolumn{1}{c|}{0.2916} & \multicolumn{1}{c|}{0.6452} & 0.4874 & \multicolumn{1}{c|}{0.4608} & \multicolumn{1}{c|}{0.3414} & \multicolumn{1}{c|}{0.6725} & 0.5631 & \multicolumn{1}{c|}{0.4396} & \multicolumn{1}{c|}{0.3347} & \multicolumn{1}{c|}{0.5274} & 0.6678 \\ 
\multicolumn{1}{c|}{}                         & BLIINDS-II* & \multicolumn{1}{c|}{0.3912} & \multicolumn{1}{c|}{0.3225} & \multicolumn{1}{c|}{0.6623} & 0.4455 & \multicolumn{1}{c|}{0.5610} & \multicolumn{1}{c|}{0.4397} & \multicolumn{1}{c|}{0.7201} & 0.6399 & \multicolumn{1}{c|}{0.5066} & \multicolumn{1}{c|}{0.3961} & \multicolumn{1}{c|}{0.6970} & 0.5973 & \multicolumn{1}{c|}{0.4626} & \multicolumn{1}{c|}{0.3609} & \multicolumn{1}{c|}{0.6814} & 0.5418 & \multicolumn{1}{c|}{0.5660} & \multicolumn{1}{c|}{0.4467} & \multicolumn{1}{c|}{0.7206} & 0.6983 & \multicolumn{1}{c|}{0.5491} & \multicolumn{1}{c|}{0.4361} & \multicolumn{1}{c|}{0.7180} & 0.6493 \\ 
\multicolumn{1}{c|}{}                         & BRISQUE*  & \multicolumn{1}{c|}{0.4276} & \multicolumn{1}{c|}{0.3387} & \multicolumn{1}{c|}{0.6701} & 0.4734 & \multicolumn{1}{c|}{0.4633} & \multicolumn{1}{c|}{0.3728} & \multicolumn{1}{c|}{0.6868} & 0.4872 & \multicolumn{1}{c|}{0.5033} & \multicolumn{1}{c|}{0.4033} & \multicolumn{1}{c|}{0.7025} & 0.5615 & \multicolumn{1}{c|}{0.3467} & \multicolumn{1}{c|}{0.2667} & \multicolumn{1}{c|}{0.6320} & 0.4179 & \multicolumn{1}{c|}{0.4987} & \multicolumn{1}{c|}{0.3921} & \multicolumn{1}{c|}{0.6950} & 0.5330 & \multicolumn{1}{c|}{0.5557} & \multicolumn{1}{c|}{0.4415} & \multicolumn{1}{c|}{0.7202} & 0.6247 \\ \hline
\multicolumn{1}{c|}{\multirow{14}{*}{FR-IQA}} & SSIM      & \multicolumn{1}{c|}{0.6282} & \multicolumn{1}{c|}{0.5178} & \multicolumn{1}{c|}{0.7607} & 0.7314 & \multicolumn{1}{c|}{0.6664} & \multicolumn{1}{c|}{0.5537} & \multicolumn{1}{c|}{0.7750} & 0.7639 & \multicolumn{1}{c|}{0.5884} & \multicolumn{1}{c|}{0.4781} & \multicolumn{1}{c|}{0.7381} & 0.7315 & \multicolumn{1}{c|}{0.6460} & \multicolumn{1}{c|}{0.5306} & \multicolumn{1}{c|}{0.7643} & 0.7444 & \multicolumn{1}{c|}{0.7561} & \multicolumn{1}{c|}{0.6164} & \multicolumn{1}{c|}{0.8060} & 0.8450 & \multicolumn{1}{c|}{0.6570} & \multicolumn{1}{c|}{0.5393} & \multicolumn{1}{c|}{0.7688} & 0.7632 \\ 
\multicolumn{1}{c|}{}                         & FSIM      & \multicolumn{1}{c|}{0.7141} & \multicolumn{1}{c|}{0.5895} & \multicolumn{1}{c|}{0.7952} & 0.7688 & \multicolumn{1}{c|}{0.7154} & \multicolumn{1}{c|}{0.5947} & \multicolumn{1}{c|}{0.7976} & 0.7738 & \multicolumn{1}{c|}{0.6642} & \multicolumn{1}{c|}{0.5496} & \multicolumn{1}{c|}{0.7750} & 0.7861 & \multicolumn{1}{c|}{0.6924} & \multicolumn{1}{c|}{0.5688} & \multicolumn{1}{c|}{0.7833} & 0.7800 & \multicolumn{1}{c|}{0.7903} & \multicolumn{1}{c|}{\textbf{0.6668}} & \multicolumn{1}{c|}{0.8321} & 0.8776 & \multicolumn{1}{c|}{0.7153} & \multicolumn{1}{c|}{0.5939} & \multicolumn{1}{c|}{0.7967} & 0.7973 \\ 
\multicolumn{1}{c|}{}                         & MS-SSIM   & \multicolumn{1}{c|}{\textbf{0.7923}} & \multicolumn{1}{c|}{\textbf{0.6706}} & \multicolumn{1}{c|}{\textbf{0.8357}} & \textbf{0.8613} & \multicolumn{1}{c|}{\textbf{0.7809}} & \multicolumn{1}{c|}{\textbf{0.6637}} & \multicolumn{1}{c|}{\textbf{0.8321}} & \textbf{0.8716} & \multicolumn{1}{c|}{\textbf{0.7349}} & \multicolumn{1}{c|}{\textbf{0.6070}} & \multicolumn{1}{c|}{\textbf{0.8024}} & \textbf{0.8557} & \multicolumn{1}{c|}{\textbf{0.7504}} & \multicolumn{1}{c|}{\textbf{0.6404}} & \multicolumn{1}{c|}{\textbf{0.8190}} & \textbf{0.8597} & \multicolumn{1}{c|}{\textbf{0.8208}} & \multicolumn{1}{c|}{\textbf{0.7000}} & \multicolumn{1}{c|}{\textbf{0.8488}} & \textbf{0.9163} & \multicolumn{1}{c|}{\textbf{0.7759}} & \multicolumn{1}{c|}{\textbf{0.6563}} & \multicolumn{1}{c|}{\textbf{0.8276}} & \textbf{0.8729} \\ 
\multicolumn{1}{c|}{}                         & IW-SSIM   & \multicolumn{1}{c|}{\textbf{0.7736}} & \multicolumn{1}{c|}{\textbf{0.6564}} & \multicolumn{1}{c|}{\textbf{0.8286}} & \textbf{0.8658} & \multicolumn{1}{c|}{07272}  & \multicolumn{1}{c|}{0.6063} & \multicolumn{1}{c|}{0.8048} & 0.8522 & \multicolumn{1}{c|}{0.6953} & \multicolumn{1}{c|}{0.5714} & \multicolumn{1}{c|}{0.7857} & \textbf{0.8357} & \multicolumn{1}{c|}{0.6984} & \multicolumn{1}{c|}{0.5760} & \multicolumn{1}{c|}{0.7869} & 0.8296 & \multicolumn{1}{c|}{0.7748} & \multicolumn{1}{c|}{0.6477} & \multicolumn{1}{c|}{0.8226} & \textbf{0.8866} & \multicolumn{1}{c|}{\textbf{0.7339}} & \multicolumn{1}{c|}{\textbf{0.6115}} & \multicolumn{1}{c|}{\textbf{0.8057}} & \textbf{0.8540} \\ 
\multicolumn{1}{c|}{}                         & PSNR      & \multicolumn{1}{c|}{0.4150} & \multicolumn{1}{c|}{0.3296} & \multicolumn{1}{c|}{0.6655} & 0.5268 & \multicolumn{1}{c|}{0.5389} & \multicolumn{1}{c|}{0.4270} & \multicolumn{1}{c|}{0.7119} & 0.6447 & \multicolumn{1}{c|}{0.4570} & \multicolumn{1}{c|}{0.3613} & \multicolumn{1}{c|}{0.6798} & 0.5806 & \multicolumn{1}{c|}{0.5860} & \multicolumn{1}{c|}{0.4780} & \multicolumn{1}{c|}{0.7381} & 0.6917 & \multicolumn{1}{c|}{0.6766} & \multicolumn{1}{c|}{0.5450} & \multicolumn{1}{c|}{0.7690} & 0.7660 & \multicolumn{1}{c|}{0.5347} & \multicolumn{1}{c|}{0.4282} & \multicolumn{1}{c|}{0.7129} & 0.6420 \\ 
\multicolumn{1}{c|}{}                         & MAD       & \multicolumn{1}{c|}{0.6867} & \multicolumn{1}{c|}{0.5705} & \multicolumn{1}{c|}{0.7857} & 0.7673 & \multicolumn{1}{c|}{0.7227} & \multicolumn{1}{c|}{0.6117} & \multicolumn{1}{c|}{0.8048} & 0.8073 & \multicolumn{1}{c|}{0.6793} & \multicolumn{1}{c|}{0.5616} & \multicolumn{1}{c|}{0.7798} & 0.8031 & \multicolumn{1}{c|}{0.7369} & \multicolumn{1}{c|}{0.6117} & \multicolumn{1}{c|}{0.8048} & 0.8246 & \multicolumn{1}{c|}{0.7834} & \multicolumn{1}{c|}{0.6547} & \multicolumn{1}{c|}{0.8250} & 0.8616 & \multicolumn{1}{c|}{0.7218} & \multicolumn{1}{c|}{0.6020} & \multicolumn{1}{c|}{0.8000} & 0.8128 \\  
\multicolumn{1}{c|}{}                         & VIF       & \multicolumn{1}{c|}{0.7384} & \multicolumn{1}{c|}{0.6088} & \multicolumn{1}{c|}{0.8060} & 0.7843 & \multicolumn{1}{c|}{0.6248} & \multicolumn{1}{c|}{0.5061} & \multicolumn{1}{c|}{0.7536} & 0.7455 & \multicolumn{1}{c|}{0.5909} & \multicolumn{1}{c|}{0.4640} & \multicolumn{1}{c|}{0.7321} & 0.7505 & \multicolumn{1}{c|}{0.6769} & \multicolumn{1}{c|}{0.5424} & \multicolumn{1}{c|}{0.7714} & 0.7567 & \multicolumn{1}{c|}{0.7214} & \multicolumn{1}{c|}{0.5828} & \multicolumn{1}{c|}{0.7917} & 0.8137 & \multicolumn{1}{c|}{0.6705} & \multicolumn{1}{c|}{0.5408} & \multicolumn{1}{c|}{0.7710} & 0.7701 \\ 
\multicolumn{1}{c|}{}                         & VSI       & \multicolumn{1}{c|}{0.5960} & \multicolumn{1}{c|}{0.4793} & \multicolumn{1}{c|}{0.7417} & 0.7295 & \multicolumn{1}{c|}{0.7114} & \multicolumn{1}{c|}{0.5875} & \multicolumn{1}{c|}{0.7929} & 0.7779 & \multicolumn{1}{c|}{0.7290} & \multicolumn{1}{c|}{0.5926} & \multicolumn{1}{c|}{0.7952} & 0.8044 & \multicolumn{1}{c|}{0.6706} & \multicolumn{1}{c|}{0.5473} & \multicolumn{1}{c|}{0.7726} & 0.8128 & \multicolumn{1}{c|}{0.7480} & \multicolumn{1}{c|}{0.6257} & \multicolumn{1}{c|}{0.8119} & 0.8513 & \multicolumn{1}{c|}{0.6910} & \multicolumn{1}{c|}{0.5665} & \multicolumn{1}{c|}{0.7829} & 0.7952 \\ 
\multicolumn{1}{c|}{}                         & GMSD      & \multicolumn{1}{c|}{0.5941} & \multicolumn{1}{c|}{0.4846} & \multicolumn{1}{c|}{0.7440} & 0.6587 & \multicolumn{1}{c|}{0.5801} & \multicolumn{1}{c|}{0.4589} & \multicolumn{1}{c|}{0.7298} & 0.6156 & \multicolumn{1}{c|}{0.6047} & \multicolumn{1}{c|}{0.4878} & \multicolumn{1}{c|}{0.7440} & 0.6719 & \multicolumn{1}{c|}{0.5527} & \multicolumn{1}{c|}{0.4401} & \multicolumn{1}{c|}{0.7190} & 0.6676 & \multicolumn{1}{c|}{0.7422} & \multicolumn{1}{c|}{0.6167} & \multicolumn{1}{c|}{0.8071} & 0.8014 & \multicolumn{1}{c|}{0.6148} & \multicolumn{1}{c|}{0.4976} & \multicolumn{1}{c|}{0.7488} & 0.6830 \\ 
\multicolumn{1}{c|}{}                         & UQI       & \multicolumn{1}{c|}{0.3682} & \multicolumn{1}{c|}{0.2771} & \multicolumn{1}{c|}{0.6405} & 0.4520 & \multicolumn{1}{c|}{0.5323} & \multicolumn{1}{c|}{0.4384} & \multicolumn{1}{c|}{0.7179} & 0.6610 & \multicolumn{1}{c|}{0.3949} & \multicolumn{1}{c|}{0.2895} & \multicolumn{1}{c|}{0.6440} & 0.4883 & \multicolumn{1}{c|}{0.5045} & \multicolumn{1}{c|}{0.3969} & \multicolumn{1}{c|}{0.6976} & 0.5998 & \multicolumn{1}{c|}{0.5265} & \multicolumn{1}{c|}{0.4137} & \multicolumn{1}{c|}{0.7060} & 0.6109 & \multicolumn{1}{c|}{0.4653} & \multicolumn{1}{c|}{0.3631} & \multicolumn{1}{c|}{0.6812} & 0.5624 \\  
\multicolumn{1}{c|}{}                         & IFC       & \multicolumn{1}{c|}{0.7411} & \multicolumn{1}{c|}{0.6112} & \multicolumn{1}{c|}{0.8071} & 0.7945 & \multicolumn{1}{c|}{0.6601} & \multicolumn{1}{c|}{0.5395} & \multicolumn{1}{c|}{0.7702} & 0.7597 & \multicolumn{1}{c|}{0.6008} & \multicolumn{1}{c|}{0.4760} & \multicolumn{1}{c|}{0.7381} & 0.7541 & \multicolumn{1}{c|}{0.6730} & \multicolumn{1}{c|}{0.5401} & \multicolumn{1}{c|}{0.7702} & 0.7589 & \multicolumn{1}{c|}{0.7214} & \multicolumn{1}{c|}{0.5805} & \multicolumn{1}{c|}{0.7905} & 0.8179 & \multicolumn{1}{c|}{0.6793} & \multicolumn{1}{c|}{0.5494} & \multicolumn{1}{c|}{0.7752} & 0.7770 \\ 
\multicolumn{1}{c|}{}                         & RFSIM     & \multicolumn{1}{c|}{0.5878} & \multicolumn{1}{c|}{0.4797} & \multicolumn{1}{c|}{0.7405} & 0.5645 & \multicolumn{1}{c|}{0.6485} & \multicolumn{1}{c|}{0.5249} & \multicolumn{1}{c|}{0.7619} & 0.6080 & \multicolumn{1}{c|}{0.5606} & \multicolumn{1}{c|}{0.4495} & \multicolumn{1}{c|}{0.7238} & 0.5468 & \multicolumn{1}{c|}{0.6932} & \multicolumn{1}{c|}{0.5520} & \multicolumn{1}{c|}{0.7750} & 0.6538 & \multicolumn{1}{c|}{0.7802} & \multicolumn{1}{c|}{0.6546} & \multicolumn{1}{c|}{0.8238} & 0.7234 & \multicolumn{1}{c|}{0.6541} & \multicolumn{1}{c|}{0.5321} & \multicolumn{1}{c|}{0.7650} & 0.6193 \\ 
\multicolumn{1}{c|}{}                         & DISTS     & \multicolumn{1}{c|}{0.6323} & \multicolumn{1}{c|}{0.5084} & \multicolumn{1}{c|}{0.7548} & 0.7409 & \multicolumn{1}{c|}{\textbf{0.7773}} & \multicolumn{1}{c|}{\textbf{0.6611}} & \multicolumn{1}{c|}{\textbf{0.8298}} & \textbf{0.8765} & \multicolumn{1}{c|}{0.6389} & \multicolumn{1}{c|}{0.5020} & \multicolumn{1}{c|}{0.7500} & 0.7939 & \multicolumn{1}{c|}{0.6734} & \multicolumn{1}{c|}{0.5380} & \multicolumn{1}{c|}{0.7679} & 0.8265 & \multicolumn{1}{c|}{0.7795} & \multicolumn{1}{c|}{0.6549} & \multicolumn{1}{c|}{0.8262} & 0.8695 & \multicolumn{1}{c|}{0.7003} & \multicolumn{1}{c|}{0.5729} & \multicolumn{1}{c|}{0.7857} & 0.8215 \\ 
\multicolumn{1}{c|}{}                         & LPIPS     & \multicolumn{1}{c|}{0.6625} & \multicolumn{1}{c|}{0.5383} & \multicolumn{1}{c|}{0.7714} & 0.7600 & \multicolumn{1}{c|}{0.7166} & \multicolumn{1}{c|}{0.5863} & \multicolumn{1}{c|}{0.7940} & 0.7954 & \multicolumn{1}{c|}{\textbf{0.7325}} & \multicolumn{1}{c|}{\textbf{0.6305}} & \multicolumn{1}{c|}{\textbf{0.8143}} & 0.7931 & \multicolumn{1}{c|}{0.6728} & \multicolumn{1}{c|}{0.5463} & \multicolumn{1}{c|}{0.7714} & 0.7792 & \multicolumn{1}{c|}{\textbf{0.7935}} & \multicolumn{1}{c|}{0.6633} & \multicolumn{1}{c|}{\textbf{0.8333}} & 0.8721 & \multicolumn{1}{c|}{0.7156} & \multicolumn{1}{c|}{0.5929} & \multicolumn{1}{c|}{0.7969} & 0.7999 \\ \hline
\multicolumn{1}{c|}{AST-IQA}                         & $Q_{\mathrm{content}}$  & \multicolumn{1}{c|}{\textbf{0.7871}} & \multicolumn{1}{c|}{\textbf{0.6707}} & \multicolumn{1}{c|}{\textbf{0.8357}} & \textbf{0.8280} & \multicolumn{1}{c|}{\textbf{0.7953}} & \multicolumn{1}{c|}{\textbf{0.6973}} & \multicolumn{1}{c|}{\textbf{0.8464}} & \textbf{0.8788} & \multicolumn{1}{c|}{\textbf{0.7528}} & \multicolumn{1}{c|}{\textbf{0.6285}} & \multicolumn{1}{c|}{\textbf{0.8131}} & \textbf{0.8507} & \multicolumn{1}{c|}{\textbf{0.7750}} & \multicolumn{1}{c|}{\textbf{0.6617}} & \multicolumn{1}{c|}{\textbf{0.8298}} & \textbf{0.8517} & \multicolumn{1}{c|}{\textbf{0.8502}} & \multicolumn{1}{c|}{\textbf{0.7454}} & \multicolumn{1}{c|}{\textbf{0.8714}} & \textbf{0.9084} & \multicolumn{1}{c|}{\textbf{0.7921}} & \multicolumn{1}{c|}{\textbf{0.6807}} & \multicolumn{1}{c|}{\textbf{0.8393}} & \textbf{0.8635} \\ \hline\hline
\end{tabular}
}
\end{center}
\end{table*}
\begin{table*}[t]
\renewcommand{\arraystretch}{1} 
\caption{Performance comparison on SR evaluation. ‘*’ indicates that the method is re-trained on the AST-IQAD.}
\label{tab5}
\begin{center}
\resizebox{2\columnwidth}{!}{
\begin{tabular}{c|ccccc|ccccccccc|c}
\hline\hline
\multicolumn{1}{c|}{\multirow{2}*{\diagbox[innerwidth=2cm]{Criteria}{Methods}}}& \multicolumn{5}{c|}{NR-IQA}                                                                                                          & \multicolumn{9}{c|}{FR-IQA}                                                                                                                                                                                                                              & AST-IQA \\ \cline{2-16} 
                        & \multicolumn{1}{c|}{NIQE}   & \multicolumn{1}{c|}{TCLT}   & \multicolumn{1}{c|}{BMPRI*} & \multicolumn{1}{c|}{BLIINDS-II*} & BRISQUE* & \multicolumn{1}{c|}{RFSIM}  & \multicolumn{1}{c|}{IFC}    & \multicolumn{1}{c|}{MAD}    & \multicolumn{1}{c|}{VIF}    & \multicolumn{1}{c|}{SSIM}   & \multicolumn{1}{c|}{IW-SSIM} & \multicolumn{1}{c|}{MS-SSIM} & \multicolumn{1}{c|}{LPIPS}  & DISTS  & $Q_{\mathrm{style}}$  \\ \hline
SRCC                    & \multicolumn{1}{c|}{0.3845} & \multicolumn{1}{c|}{0.2019} & \multicolumn{1}{c|}{\textbf{0.4479}} & \multicolumn{1}{c|}{0.4162}     & \textbf{0.4885}   & \multicolumn{1}{c|}{0.0016} & \multicolumn{1}{c|}{0.1008} & \multicolumn{1}{c|}{0.2769} & \multicolumn{1}{c|}{0.1744} & \multicolumn{1}{c|}{0.3093} & \multicolumn{1}{c|}{0.1243}  & \multicolumn{1}{c|}{0.3252}  & \multicolumn{1}{c|}{0.2471} & 0.4233 & \textbf{0.6062}  \\ 
KRCC                    & \multicolumn{1}{c|}{0.3075} & \multicolumn{1}{c|}{0.1396} & \multicolumn{1}{c|}{\textbf{0.3483}} & \multicolumn{1}{c|}{0.3301}     & \textbf{0.3952}   & \multicolumn{1}{c|}{0.0001} & \multicolumn{1}{c|}{0.0725} & \multicolumn{1}{c|}{0.2146} & \multicolumn{1}{c|}{0.1196} & \multicolumn{1}{c|}{0.2300} & \multicolumn{1}{c|}{0.0978}  & \multicolumn{1}{c|}{0.2425}  & \multicolumn{1}{c|}{0.1955} & 0.3327 & \textbf{0.4886}  \\ 
HITR                    & \multicolumn{1}{c|}{0.6538} & \multicolumn{1}{c|}{0.5702} & \multicolumn{1}{c|}{\textbf{0.6744}} & \multicolumn{1}{c|}{0.6641}     & \textbf{0.6975}   & \multicolumn{1}{c|}{0.5002} & \multicolumn{1}{c|}{0.5360} & \multicolumn{1}{c|}{0.6076} & \multicolumn{1}{c|}{0.5590} & \multicolumn{1}{c|}{0.6145} & \multicolumn{1}{c|}{0.5469}  & \multicolumn{1}{c|}{0.6202}  & \multicolumn{1}{c|}{0.5971} & 0.6631 & \textbf{0.7412}  \\ 
PLCC                    & \multicolumn{1}{c|}{0.4295} & \multicolumn{1}{c|}{0.2183} & \multicolumn{1}{c|}{\textbf{0.5169}} & \multicolumn{1}{c|}{0.4651}     & \textbf{0.5254}   & \multicolumn{1}{c|}{0.0135} & \multicolumn{1}{c|}{0.0890} & \multicolumn{1}{c|}{0.3389} & \multicolumn{1}{c|}{0.1634} & \multicolumn{1}{c|}{0.3270} & \multicolumn{1}{c|}{0.1314}  & \multicolumn{1}{c|}{0.3925}  & \multicolumn{1}{c|}{0.2608} & 0.4654 & \textbf{0.6278} \\ \hline\hline
\end{tabular}
}
\end{center}
\end{table*}

\emph{1) Performance Test on CP:} Although the evaluation of CP between the stylized images and source content images is not a classic FR-IQA problem, since the stylized image targets to maintain the structure information of source content image, the structure measurement module commonly included in existing FR-IQA methods is relatively suitable for comparison. As a consequence, we compare the proposed $Q_{\mathrm{content}}$ with the state-of-the-art general-purpose FR-IQA metrics. In addition, we also utilize the NR-IQA method to establish the functional mapping from the stylized images to the CP quality scores. Each content subset conducts the same training-testing strategies described above. The performance compassion results are listed in Table \ref{tab4}, where the top three metrics are highlighted in bold. It can be seen that the traditional general-purpose FR-IQA metrics perform better than NR-IQA metric. It is reasonable since the purpose of CP evaluation is to measure the content structure similarity between the stylized and the original content images. Thus, ignoring the content image and directly extracting features from the stylized image for regression not only lacks enough useful information, but also has no practical significance. Our $Q_{\mathrm{content}}$ achieves the best performance for all dataset on the three most important ranking-related performance criteria: SRCC, KRCC and HITR, but is inferior to MS-SSIM \cite{ref57} on PLCC. The reason is that MS-SSIM \cite{ref57} also applies the multi-scale feature extraction strategy to simulate the visual characteristics of humans appreciating art works from different scales.

\emph{2) Performance Test on SR:} To our best knowledge, there is no related methods for evaluating the quality of stylized images on SR. Although SR evaluation cannot be taken as a classical FR/NR-IQA issue, we are curious about how the performance of these methods in AST-IQA task, especially given the lack of comparison methods. Thus, we report the performance comparison results in Table \ref{tab5}. It is clear that the above FR-IQA methods are not suitable for quantitative evaluation of SR, because of the difference in contents between the stylized images and the style images. DISTS \cite{ref46} obtains the better performance among these FR-IQA metrics, since it is designed to evaluate structural and texture (related to style) similarities, allowing for slight pixel misalignment. In addition, although the NR-IQA method can map the stylized images to SR quality scores with powerful learning machines, it lacks practical relevance. In addition, it is also doubtful whether the NR-IQA method can maintain high performance when testing more diverse images that are not included in training. Compared with other methods, our $Q_{\mathrm{style}}$ achieves the best performance, since it builds a strong association (e.g., style pattern and brush stoke) with AST.
\begin{table}[t]
\renewcommand{\arraystretch}{1.1} 
\caption{Performance comparison on OV evaluation. ‘*’ indicates that the method is re-trained on the AST-IQAD. The subscript "SR/CP" of FR-IQA methods represent the quality score generating from the SR/CP evaluation.}
\begin{center}
\label{tab6}
\resizebox{1\columnwidth}{!}{
\begin{tabular}{c|c|c|c|c|c}
\hline\hline
Type                     & Method        & SRCC   & KRCC   & HITR   & PLCC   \\ \hline
\multirow{9}{*}{NR-IQA}  & NIQE          & 0.2615 & 0.2078 & 0.6036 & 0.3041 \\ 
                         & TCLT          & 0.0189 & 0.0177 & 0.5081 & 0.0601 \\ 
                         & BMPRI*        & 0.3705 & 0.2832 & 0.6413 & 0.4739 \\  
                         & BLIINDS-II*    & 0.3259 & 0.2467 & 0.6228 & 0.4201 \\  
                         & BRISQUE*      & 0.4124 & 0.3179 & 0.6577 & 0.4685 \\ 
                         & UNIQUE        & 0.2038 & 0.1507 & 0.5776 & 0.3063 \\  
                         & WaDIQaM*      & 0.3779 & 0.2840 & 0.6421 & 0.4183 \\ 
                         & TReS*         & \textbf{0.5993} & \textbf{0.4816} & \textbf{0.7398} & \textbf{0.6779} \\ \hline
\multirow{18}{*}{FR-IQA} & $\mathrm{UQI_{CP}}$         & 0.3252 & 0.2365 & 0.6176 & 0.3626 \\  
                         & $\mathrm{IFC_{CP}}$         & 0.3976 & 0.2955 & 0.6483 & 0.5015 \\  
                         & $\mathrm{VIF_{CP}}$         & 0.3970 & 0.2927 & 0.6469 & 0.5032 \\ 
                         & $\mathrm{VSI_{CP}}$         & 0.3714 & 0.2744 & 0.6369 & 0.4477 \\  
                         & $\mathrm{PSNR_{CP}}$        & 0.3275 & 0.2386 & 0.6188 & 0.3791 \\  
                         & $\mathrm{MAD_{CP}}$         & 0.3482 & 0.2501 & 0.6250 & 0.4122 \\  
                         & $\mathrm{SSIM_{CP}}$         & 0.3495 & 0.2530 & 0.6267 & 0.4268 \\  
                         & $\mathrm{SSIM_{SR}}$         & 0.1890 & 0.1512 & 0.5757 & 0.1930 \\  
                         & $\mathrm{FSIM_{CP}}$         & 0.3749 & 0.2773 & 0.6388 & 0.4399 \\  
                         & $\mathrm{GMSD_{CP}}$        & 0.2814 & 0.2062 & 0.6033 & 0.3250 \\ 
                         & $\mathrm{RFSIM_{CP}}$        & 0.3585 & 0.2616 & 0.6300 & 0.2968 \\  
                         & $\mathrm{RFSIM_{SR}}$        & 0.0254 & 0.0209 & 0.5095 & 0.0204 \\ 
                         & $\mathrm{DISTS_{CP}}$        & 0.3626 & 0.2681 & 0.6340 & 0.4875 \\ 
                         & $\mathrm{DISTS_{SR}}$        &\textbf{0.4695} & \textbf{0.3693} & \textbf{0.6829} & \textbf{0.5169} \\ 
                         & MS-$\mathrm{SSIM_{CP}}$      & 0.4258 & 0.3201 & 0.6598 & 0.5174 \\  
                         & MS-$\mathrm{SSIM_{SR}}$      & 0.2625 & 0.2063 & 0.6033 & 0.3334 \\  
                         & IW-$\mathrm{SSIM_{CP}}$      & 0.4177 & 0.3089 & 0.6550 & 0.5258 \\  
                         & IW-$\mathrm{SSIM_{SR}}$      & 0.1238 & 0.0988 & 0.5486 & 0.1285 \\ \hline
\multirow{3}{*}{AST-IQA} & $Q_{\mathrm{content}}$      & 0.4520 & 0.3439 & 0.6724 & 0.5163 \\ 
                         & $Q_{\mathrm{style}}$        & 0.2313 & 0.1779 & 0.5886 & 0.2310 \\  
                         & $Q_{\mathrm{overall}}$      &\textbf{ 0.6077} & \textbf{0.4855} & \textbf{0.7410} & \textbf{0.6510} \\ \hline\hline
\end{tabular}
}
\end{center}
\end{table}

\emph{3) Performance Test on OV:} Since there is no ground truth for stylized image to directly compare on OV evaluation, instead of using the FR-IQA methods, we utilize the CP/SR quality score of the FR-IQA method for performance comparison, which is beneficial to analyze and explore the contribution of the CP/SR quality components on the overall quality OV. Additionally, we also perform NR-IQA on the OV quality evaluation task, which will be useful to understand how challenging this task is for the existing NR-IQA metrics. The performance comparison results of all methods are shown in Table \ref{tab6}. In the table, the subscript "SR/CP" represents the quality scores generating from the SR/CP evaluation. From the results, we observe that: 1) All CP and SR quality score generated by vanilla FR metrics have weak correlation with the overall quality OV. It indicates that OV quality evaluation is a complex process where multiple factors need to be considered. 2) The learning-based NR-IQA methods show good performance results, in particular, transformer-based TReS \cite{ref68} achieves competitive results with our $Q_{\mathrm{overall}}$. Obviously, utilizing deep neural network to perform AST-IQAD task is a promising way. 3) It can be seen that $Q_{\mathrm{overall}}$ has a higher performance than $Q_{\mathrm{style}}$ or $Q_{\mathrm{content}}$ alone, which also demonstrates the effectiveness of our pooling strategy. As an unsupervised learning method based on sparse representation, our $Q_{\mathrm{overall}}$ can stably evaluate OV quality scores and achieve the best performance on three criteria via properly combining the $Q_{\mathrm{style}}$ and $Q_{\mathrm{content}}$. Actually, as discussed in Section IV-C, there still leaves a large space for improving the evaluation accuracy via proper importance weights and combination strategies.

\begin{figure}[!t]
    \centering
    \includegraphics[width=3.5in]{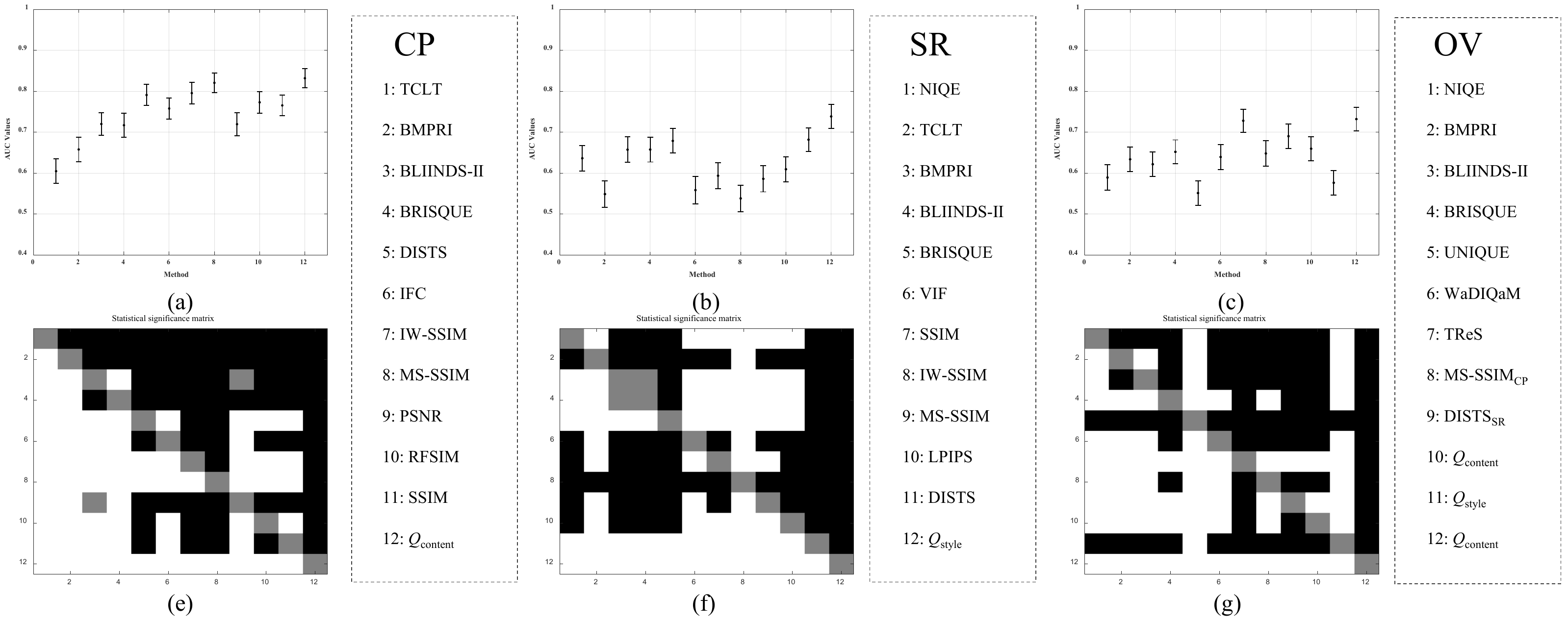}
    \caption{Statistical significance analyses. (a)-(c): AUC values of different method on CP, SR, and OV evaluation, respectively. (e)-(f): two-sample t-test results (statistical significance matrix) on CP, SR, and OV evaluation, respectively.}
    \label{fig17}
\end{figure}
\begin{figure}[!t]
    \centering
    \includegraphics[width=3.5in]{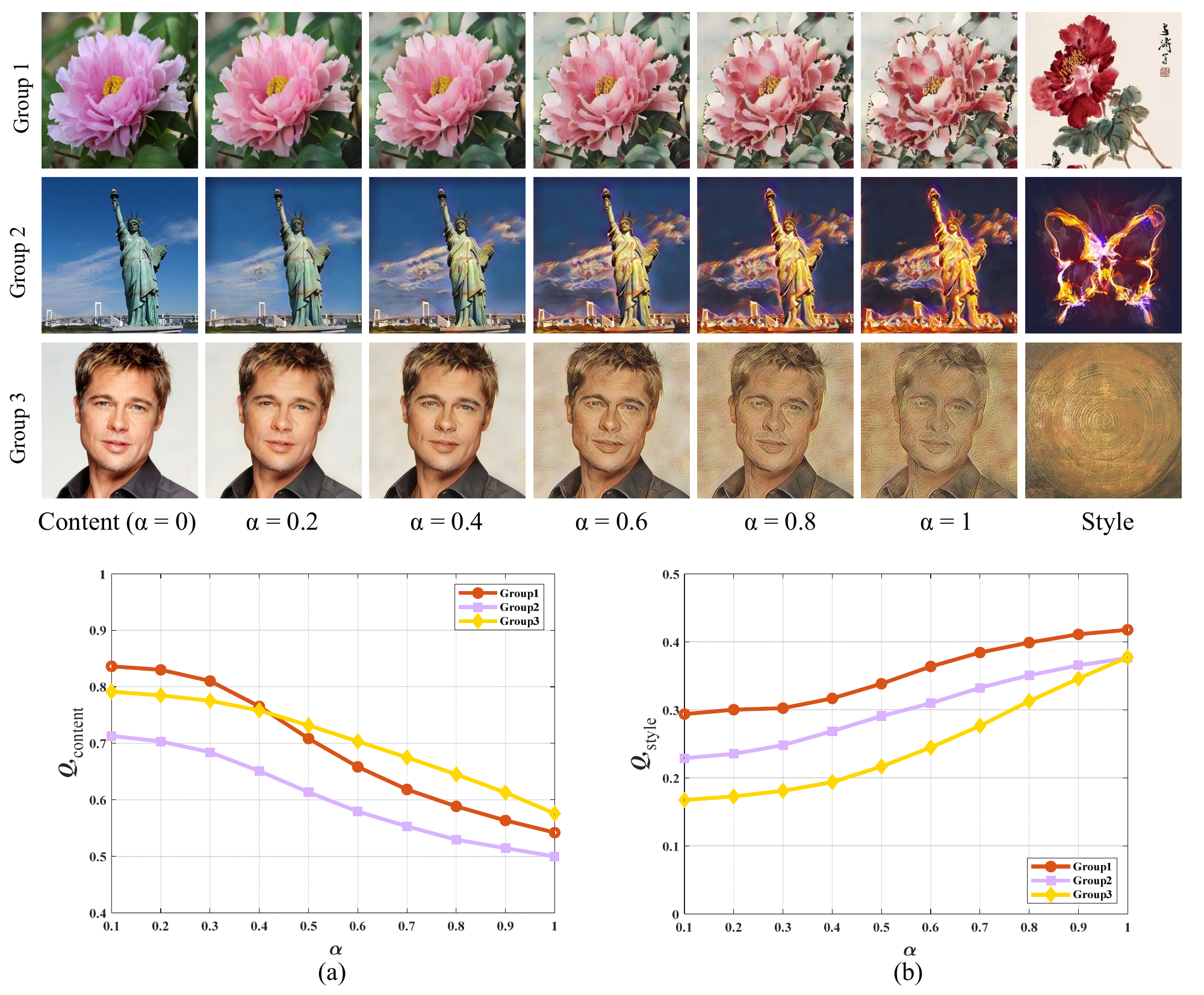}
    \caption{Robustness analysis of the proposed $Q_{\mathrm{style}}$ and $Q_{\mathrm{content}}$ in content and style trade-off application.}
    \label{fig18}
\end{figure}
\begin{figure}[!t]
    \centering
    \includegraphics[width=3.3in]{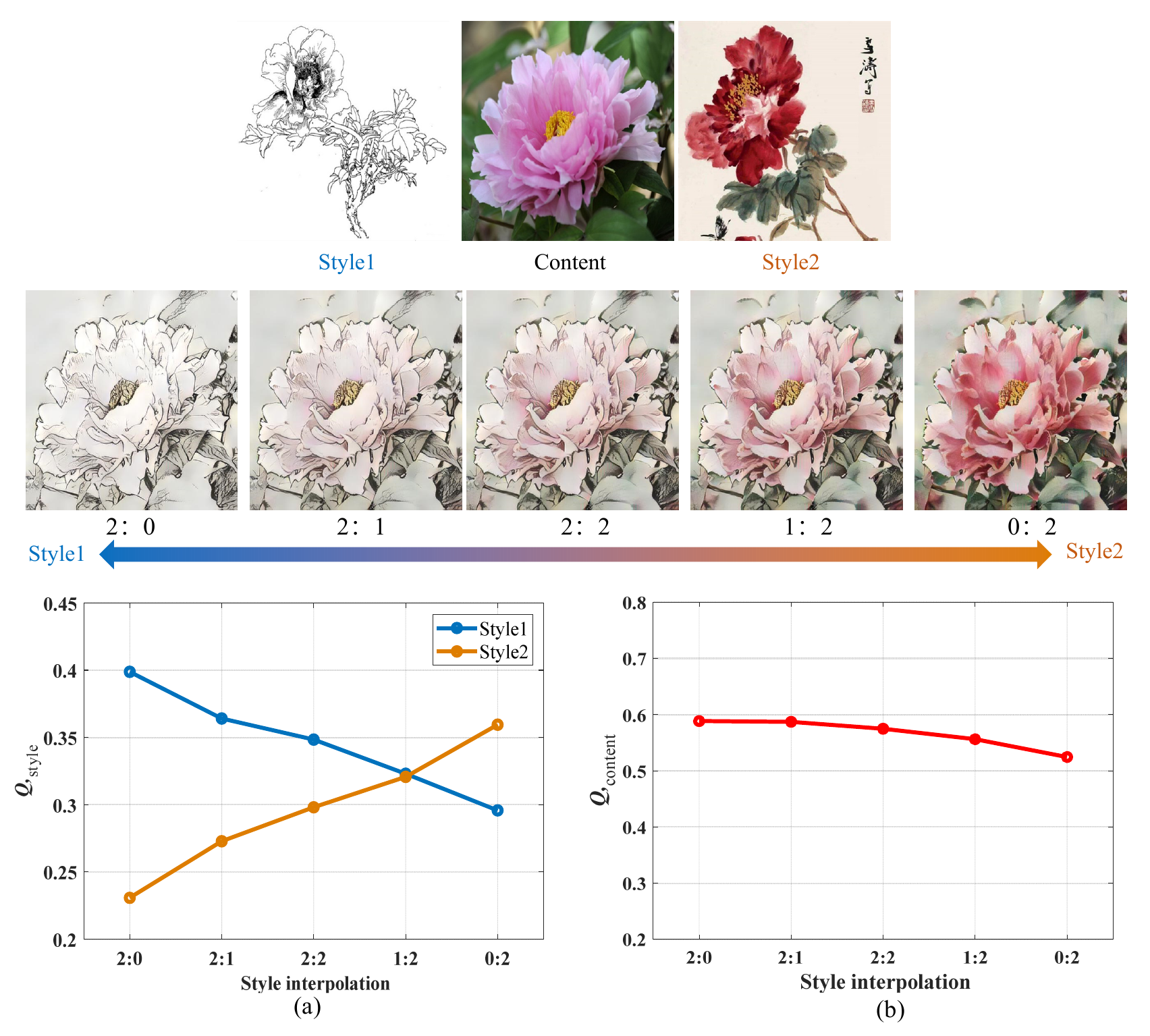}
    \caption{Robustness analysis of the proposed $Q_{\mathrm{style}}$ and $Q_{\mathrm{content}}$ in style interpolation application.}
    \label{fig19}
\end{figure}
\subsection{Statistical Analysis}
In the above experiments, the proposed SROE demonstrates a better correlation between the prediction scores and ground truth. We further adopt the hypothesis testing approach based on t-statistics \cite{ref69} to prove that our SRQE is statistically better than other metrics. Specifically, we first calculate the area under curve (AUC) values (i.e., the area covered by receiver operating characteristic (ROC)) with 95\% confidence interval (CI) for all image pairs (from PC in the subjective study). A higher value of AUC indicates better performance of the method. Next, we carry out the two-sample t-test between the pair of AUC values with 95\% CI. We show the results of the AUC values and statistical significance of difference in Fig. \ref{fig17}, where the white/gray/black square manifests that the method in row is significantly better/indistinguishable/worse than the method in column. It can remark from the results that our SRQE	performs significantly better than all competitors, indicating the superiority of our SRQE method.

\subsection{Robustness Analysis}
In this subsection, we present the robustness of our proposed $Q_{\mathrm{style}}$ and $Q_{\mathrm{content}}$ in two AST applications (i.e., content-style trade-off and style interpolation) which are included in many AST methods.

\emph{1) Content-style trade-off:} This application can adjust the degree of stylization. Three style degree groups with smooth changes generated by MANet \cite{ref17} are presented in Fig. \ref{fig18}. When $\alpha$ = 1, the fully stylized image is obtained. Fig. \ref{fig18} (a)-(b) present $Q_{\mathrm{content}}$ and $Q_{\mathrm{style}}$ results with different degrees of stylization. It can be seen that as $\alpha$ increases from 0 to 1, the $Q_{\mathrm{content}}$ (or $Q_{\mathrm{style}}$) value is consequently decreasing (or increasing) gradually which indicates that our $Q_{\mathrm{style}}$ and $Q_{\mathrm{content}}$ can effectively capture the changes in the style patterns and content structure of the image.

\emph{2) Style interpolation:} This application is to merge multiple style images into a single generated result. Here, we also utilize MANet \cite{ref17} to generate a group of stylized images with different interpolations, and then use $Q_{\mathrm{style}}$ (or $Q_{\mathrm{content}}$) to evaluate SR (or CP) of the stylized images. As shown in the Fig. \ref{fig19} (a), with the continuous decline of the weights for the specific styles, the $Q_{\mathrm{style}}$ value is consequently decreasing gradually which indicates that our $Q_{\mathrm{style}}$ can clarify style characteristics and accurately evaluate SR even under the interference of multiple styles. Fig. \ref{fig19} (b) presents that our $Q_{\mathrm{content}}$ predicts a set of slowly decreasing quality values as the style changes from Style1 (Line drawing) to Style2 (Xieyi). It is reasonable because the Line drawing pays more attention to the maintenance of structure and shape than Xieyi. The results of the above predictions prove the robustness of our method.
\begin{figure}[!t]
    \centering
    \includegraphics[width=3.3in]{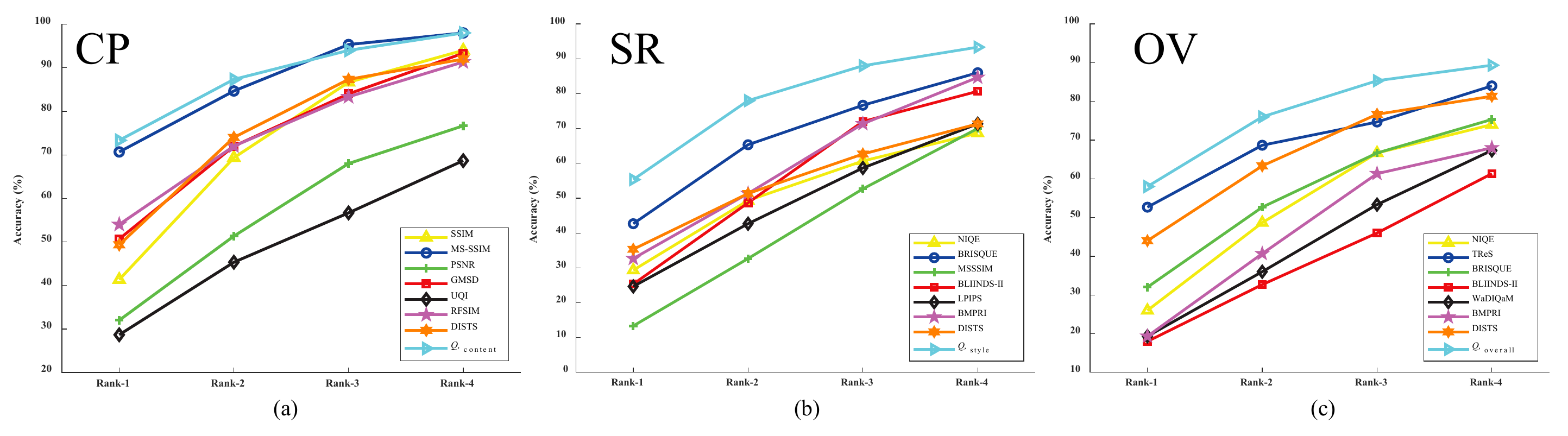}
    \caption{Rank-n accuracy on the three quality factors.}
    \label{fig20}
\end{figure}
\begin{figure}[!t]
    \centering
    \includegraphics[width=3.3in]{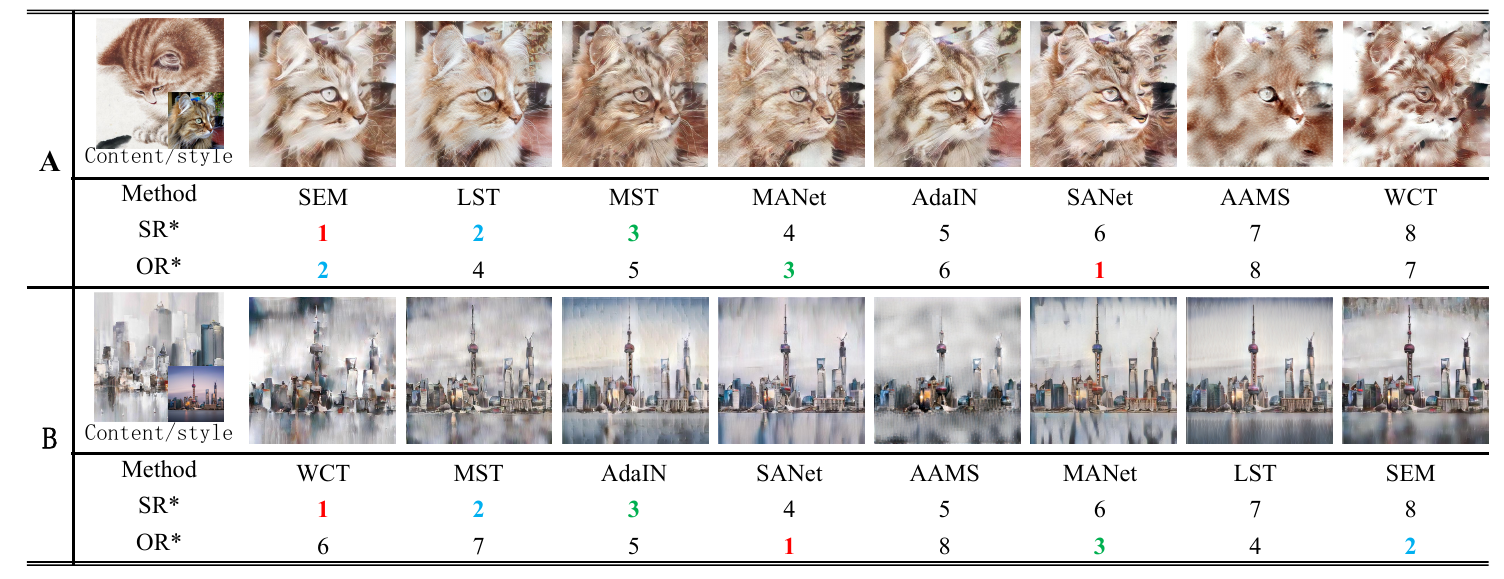}
    \caption{Some typical failure ranking of our method on the OV evaluation. SR* denotes the subjective ranking, and OR* denotes the objective ranking.}
    \label{fig21}
\end{figure}
\subsection{Ranking Capability}
\emph{1) Performance Test on Rank-n accuracy:} To comprehensively compare the performance of the IQA metrics, we also focus on the performance of Rank-n accuracy \cite{ref70}. Given the eight stylized images for each Content-Style image pair, the rank-n accuracy is the percentage of the objective scores where the subjective-rated best one is within their top n positions. The results are presented in Fig. \ref{fig20}. We can observe that the proposed method achieves a fairly good performance in the SR, CP and OV ranking accuracy tests. Since one of the most important applications of AST-IQA metric is to guide the generation of stylized images, the proposed method is a promising tool for automatic selection of the optimal transferring result from a set of candidates.

\emph{2) Failure Ranking Analysis:} As aforementioned, we demonstrate the ranking capability of the proposed method on the three quality factors. However, in some special situations, the proposed method encounters challenges to achieve the expected ranking results. As shown in Fig.  \ref{fig21}, we present two groups of representative failure rankings on the OV evaluation, which can be divided into two categories. In the first category, our method fails to capture extraneous artifacts, as shown in Fig. \ref{fig21} (A). We select the stylization produced by SANet \cite{ref16} as the optimal result, which produces unpleasing eye-like artifacts (zoom in for greater clarity) due to the fine-grained nature. The reason behind this lies in that our method mainly uses global sparse representation and ignores the local extraneous artifacts. In the second category, our method fails to effectively balance the importance of quality factors, as show in Fig. \ref{fig21} (B). Actually, the CP and SR do not always complement each other. A non-realistic style leads to lower content retention in the final stylization result. Obviously, there is a large room to manipulate the importance weights for the quality factors. Overall, how to further dig the aesthetic information behind the stylization and propose a better strategy to balance different quality factors are the key issues to be explored in the future work.

\section{Conclusion}
In this paper, we first constructed a new database (AST-IQAD) to collect the subject-rated scores on the three quality factors of content preservation (CP), style resemblance (SR), and overall vision (OV). Then, a new sparse representation-based method (SRQE) is proposed to predict the human perception toward different stylized results. Experimental results show that our proposed method produces very promising AST-IQA results compared with existing general-purpose IQA methods. Overall, our new database creates a reliable platform to evaluate the performance of different AST algorithms and our method is helpful for guiding the design of different algorithms. In our future work, we will further mine the aesthetic information behind the stylization and propose a better strategy to balance different quality factors.

\begin{IEEEbiography}[{\includegraphics[width=1in,height=1.25in,clip,keepaspectratio]{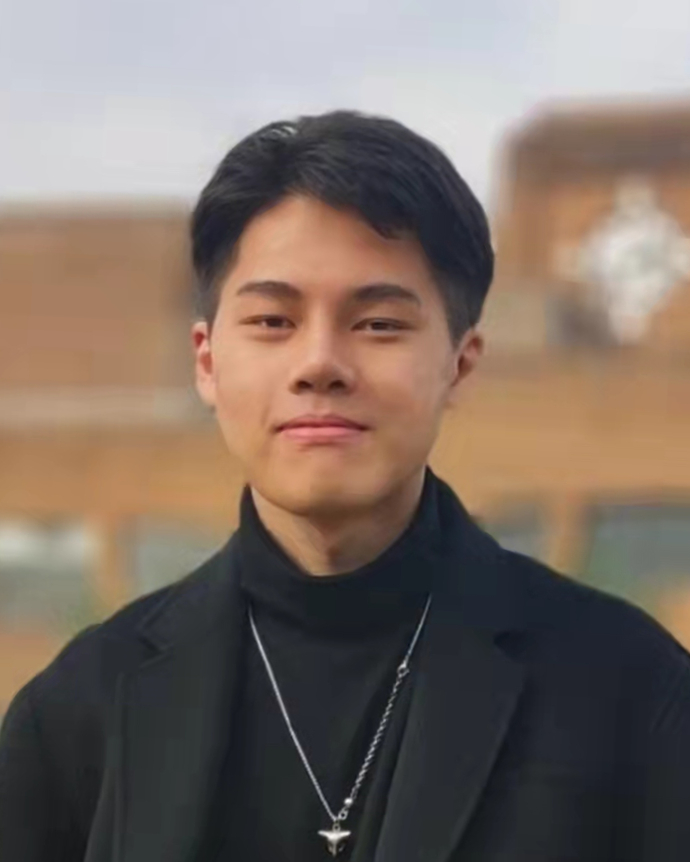}}]{Hangwei Chen}
received the B.S. degree from Ningbo University, China, in 2020. He is currently pursuing the Ph.D. degree in Signal and Information Processing at Ningbo University, Ningbo,
China. His current research interests include image processing and quality assessment.
\end{IEEEbiography}

\begin{IEEEbiography}[{\includegraphics[width=1in,height=1.25in,clip,keepaspectratio]{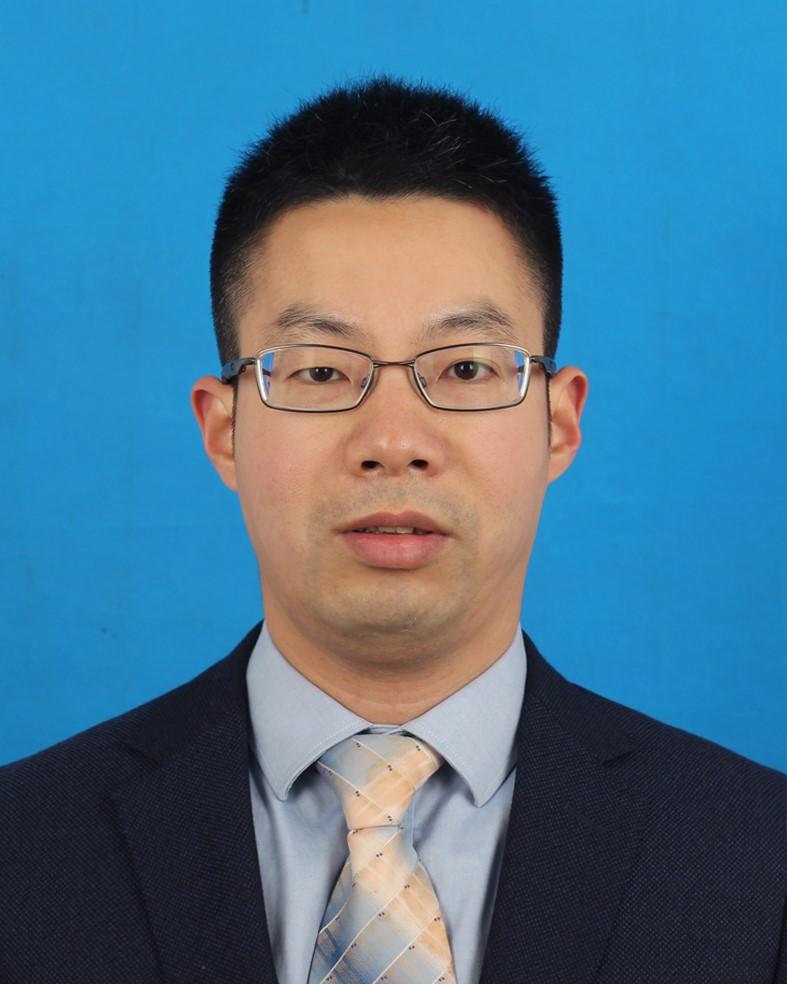}}]{Feng Shao}
 (M’16) received his B.S. and Ph.D degrees from Zhejiang University, Hangzhou, China, in 2002 and 2007, respectively, all in Electronic Science and Technology. He is currently a professor in Faculty of Information Science and Engineering, Ningbo University, China. He was a visiting Fellow with the School of Computer Engineering, Nanyang Technological University,
Singapore, from February 2012 to August 2012. He received ‘Excellent
Young Scholar’ Award by NSF of China (NSFC) in 2016. He has published
over 100 technical articles in refereed journals and proceedings in the areas of
3D video coding, 3D quality assessment, and image perception, etc.
\end{IEEEbiography}

\begin{IEEEbiography}[{\includegraphics[width=1in,height=1.25in,clip,keepaspectratio]{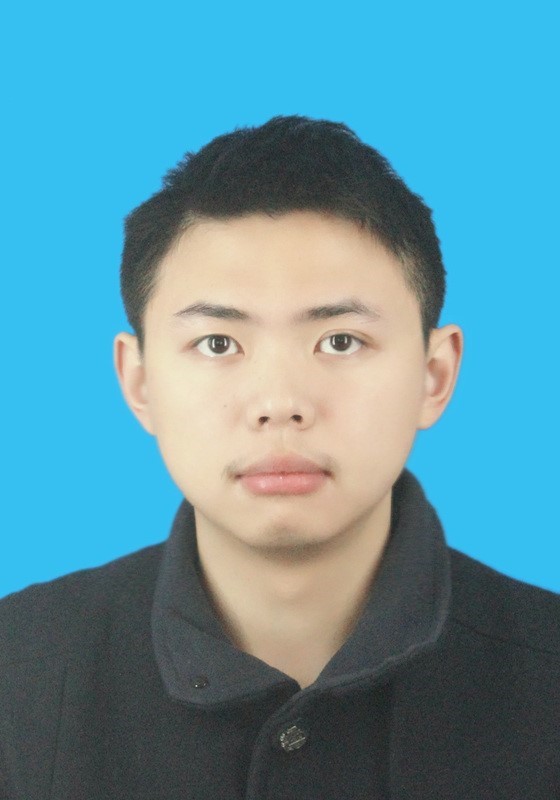}}]{Xiongli Chai}received the B.S. degree and M.S. degree from Ningbo University, China, in 2017 and 2020 respectively. He is currently pursuing the Ph.D. degree in Signal and Information Processing at Ningbo University, Ningbo, China. His current
research interests include image/video processing and quality assessment. 
\end{IEEEbiography}

\begin{IEEEbiography}[{\includegraphics[width=1in,height=1.25in,clip,keepaspectratio]{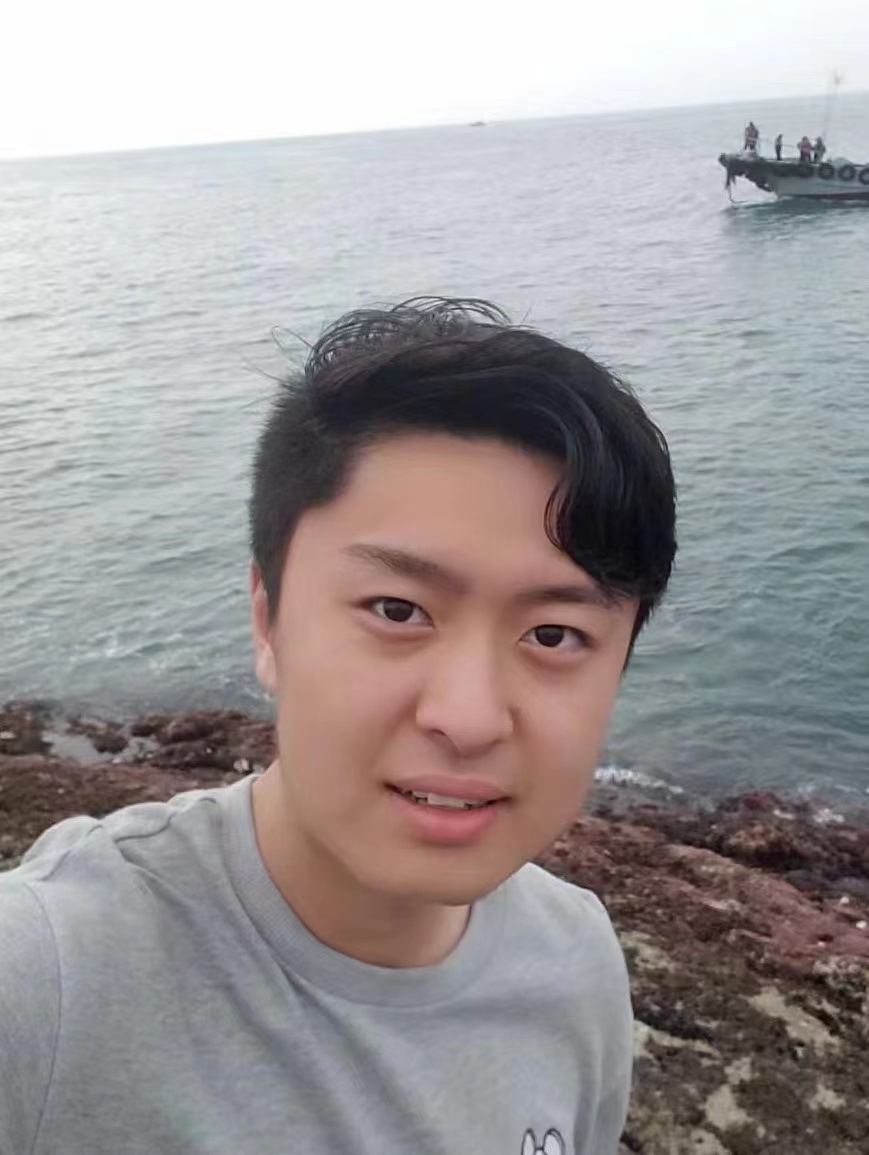}}]{Yuese Gu} received the B.S. degree in communication engineering from Ningbo University, Ningbo, China, where he is currently pursuing the master’s degree. His research interest lies in deep learning with applications in image processing and computer vision.
\end{IEEEbiography}

\begin{IEEEbiography}[{\includegraphics[width=1in,height=1.25in,clip,keepaspectratio]{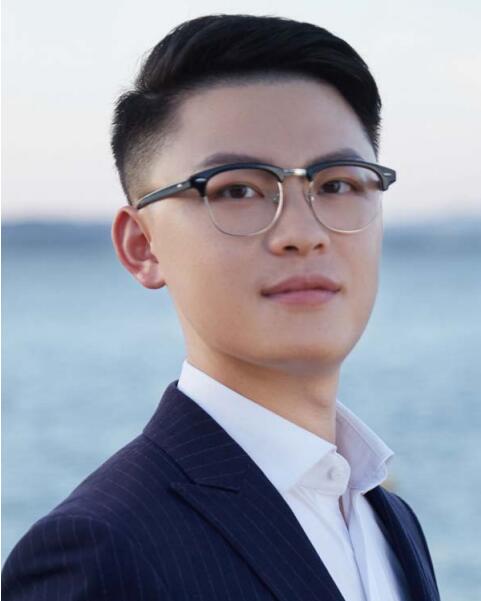}}]{Qiuping Jiang} (M’18) received the Ph.D. degree in signal and information processing from Ningbo University in 2018. From January 2017 to May 2018, he was a Visiting Student with Nanyang Technological University, Singapore. He is currently an Associate Professor with Ningbo University. His research interests include image processing, visual perception, and computer vision. He has received the 2017 Best Paper Honorable Mention Award of the Journal of Visual Communication and Image Representation. He serves as an Associate Editor for IET Image Processing, Journal of Electronic Imaging, and APSIPA Transactions on Signal and Information Processing.
\end{IEEEbiography}

\begin{IEEEbiography}[{\includegraphics[width=1in,height=1.25in,clip,keepaspectratio]{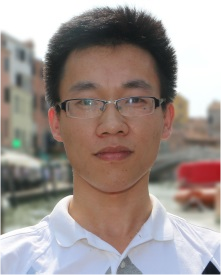}}]{Xiangchao Meng}
 (M’18)  received the B.S. degree in geographic information system from the Shandong University of Science and Technology, Qingdao, China, in 2012, and the Ph.D. degree in cartography and geography information system from Wuhan University, Wuhan, China, in 2017. He is currently a Lecturer with the Faculty of Electrical Engineering and Computer Science, Ningbo University, Ningbo, China. His research interests include variational methods and remote sensing image fusion.
\end{IEEEbiography}

\begin{IEEEbiography}[{\includegraphics[width=1in,height=1.25in,clip,keepaspectratio]{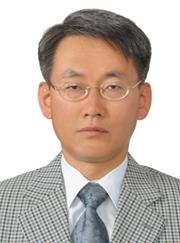}}]{Yo-Sung Ho}
 (SM’06–F’16) received the B.S. and M.S. degrees
in electronic engineering from Seoul National University, Seoul,
Korea, in 1981 and 1983, respectively, and the Ph.D. degree in
electrical and computer engineering from the University of California, Santa Barbara, in 1990. He joined Electronics and Telecommunications Research Institute (ETRI), Daejon, Korea, in 1983. From 1990 to 1993, he was with Philips Laboratories, Briarcliff Manor, NY, where
he was involved in development of the Advanced Digital High-Definition
Television (AD-HDTV) system. In 1993, he rejoined the technical staff of
ETRI and was involved in development of the Korean DBS digital television
and high-definition television systems. Since 1995, he has been with Gwangju
Institute of Science and Technology (GIST), Gwangju, Korea, where he is
currently Professor of Information and Communications Department. His
research interests include digital image and video coding, image analysis and
image restoration, advanced video coding techniques, digital video and audio
broadcasting, three-dimensional video processing, and content-based signal
representation and processing.
\end{IEEEbiography}

\end{document}